\documentclass[lettersize,journal]{IEEEtran}
\usepackage{amsmath,amsfonts}
\usepackage{algorithmic}
\usepackage{algorithm}
\usepackage{array}
\usepackage[caption=false,font=normalsize,labelfont=sf,textfont=sf]{subfig}
\usepackage{textcomp}
\usepackage{stfloats}
\usepackage{url}
\usepackage{verbatim}
\usepackage{graphicx}
\usepackage{cite}
\usepackage{multirow}
\usepackage{booktabs}
\usepackage{algorithm}
\usepackage{algorithmic}
\usepackage{color}
\usepackage{multirow}
\usepackage{amsmath}
\usepackage{amsmath, amssymb, bm}
\usepackage{xcolor}

\usepackage{booktabs} 
\usepackage{pifont}

\begin{document}

\title{WaveSR: Physics-Inspired Wave Evolution Attention for Efficient Image Super-Resolution}

\author{
Xiaoyan Lei, \IEEEmembership{Graduate Student Member, IEEE},  
Biao Luo,
\IEEEmembership{Senior Member, IEEE},
Wenlong Zhang, 
Tingwen Huang,
\IEEEmembership{Fellow, IEEE},
Chunhua Yang,
\IEEEmembership{Fellow, IEEE},
Weihua Gui,
\IEEEmembership{Member, IEEE},
}

\markboth{IEEE Transactions on Multimedia}%
{Shell \MakeLowercase{\textit{et al.}}: A Sample Article Using IEEEtran.cls for IEEE Journals}


\maketitle

\begin{abstract}
Recently, physics-inspired neural networks have gained widespread attention due to their interpretability. The closed-form, frequency-time decoupled solution of the wave equation offers linear complexity and allows different frequency components to evolve independently, making it a potential alternative solution for efficient image super-resolution. 
Therefore, a physics-inspired wave evolution attention framework is proposed, namely WaveSR. The WaveSR framework involves two key modules, i.e., wave evolution attention (WEA) and Frequency-Adaptive Feed-Forward Network (FA-FFN). 
The WEA module achieves feature interactions through the analytical solution of the 2D wave equation in the Fast Fourier Transform (FFT) domain, enabling independent evolution of different frequency components with learnable wave speed \(c\) and damping coefficient \(\alpha\).
The FA-FFN module realizes adaptive frequency feature selection through learnable spectral modulation, effectively enhancing feature representation capability.
Subsequently, we employ a two-stage training strategy with a hybrid objective that combines a spatial-domain pixel loss and a frequency-domain loss to improve reconstruction quality.
Our method provides a preliminary exploration of physics-inspired neural networks for efficient image super-resolution.

\end{abstract}


\section{Introduction}
\IEEEPARstart{I}{m}age Super-Resolution (SR), aiming to reconstruct high-resolution (HR) images from their low-resolution (LR) counterparts, is a fundamental yet challenging problem in computer vision with numerous practical applications. Super-resolution is inherently ill-posed, as a given low-resolution input may correspond to numerous plausible high-resolution outputs. 
Addressing this challenge requires effective prior knowledge and sophisticated modeling techniques to faithfully recover high-frequency details. 
Deep neural networks have revolutionized SR performance, evolving from shallow CNNs~\cite{srcnn,kim2016accurate} to increasingly complex architectures incorporating residual learning~\cite{lim2017enhanced}, attention mechanisms~\cite{cao2025hasn,liu2020residual}, and most recently, Transformers~\cite{liang2021swinir,chen2023hat}. Furthermore, research in directions such as degradation modeling~\cite{closerlookBSR}, multi-degradation task conflicts~\cite{tgsr}, and prompt learning~\cite{chen2024learning} has continuously advanced the development of the SR field. 
\begin{figure}[!tbp]
\centering
\includegraphics[width=3.3in]{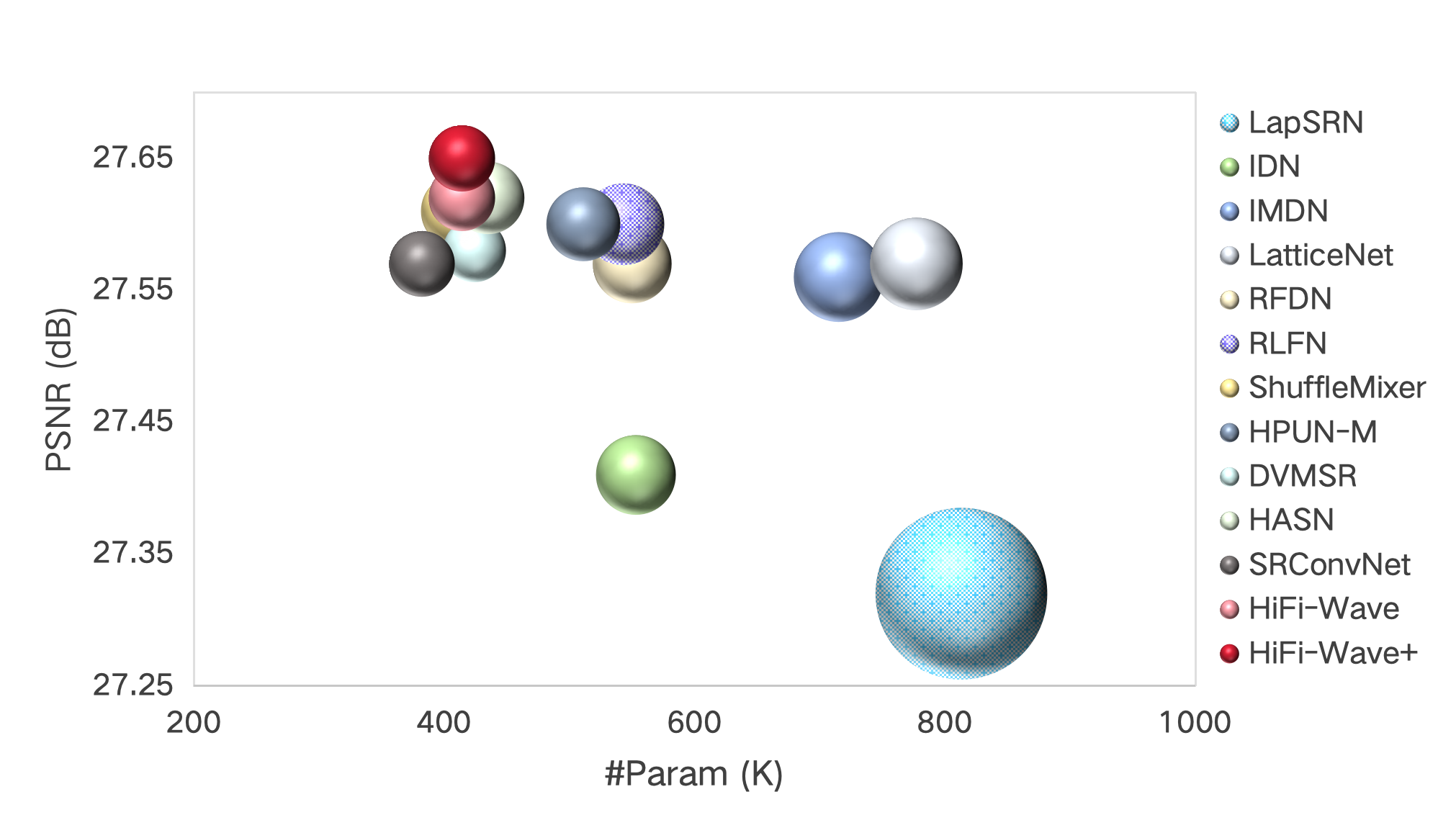}
\caption{Comparison of the proposed method with state-of-the-art approaches on the BSD100 dataset~\cite{B100} in terms of PSNR, network parameters, and FLOPs. The circle size represents the FLOPs.}
\label{fig:intro}
\end{figure}
While these advances have consistently pushed the state-of-the-art, they come at the cost of substantial computational overhead and parameter budgets. This trade-off becomes particularly acute in resource-constrained scenarios such as mobile devices and edge computing, where the demand for efficient yet high-quality SR has catalyzed research into lightweight architectures.

A common thread among existing lightweight SR methods is the pursuit of a delicate balance between model compactness and reconstruction fidelity~\cite{fsrcnn,IDN2018,hui2019lightweight,liu2025catanet,li2026fouriersr,li2024efficient,camixersr}. These methods typically focus on efficient convolution designs, stacked residual learning, and carefully crafted lightweight modules to achieve favorable efficiency performance trade-offs. Many approaches aim to separate high-frequency details from low-frequency structures, and considerable progress has been made in edge enhancement and texture recovery under resource-limited settings~\cite{du2026wavelet,wang2025lightweight}. Meanwhile, an increasing number of lightweight SR frameworks have begun to explore frequency-domain properties of image signals, leading to improved efficiency in feature computation and information propagation~\cite{kong2023efficient, xu2024fdsr, dong2025dual}. 
These data-driven approaches have substantially advanced lightweight image restoration by achieving a better trade-off between accuracy and complexity. However, most existing methods derive their feature transformation and interaction mechanisms primarily from learned architectural designs, while the potential of incorporating physics-inspired inductive biases into lightweight SR networks remains relatively underexplored.

Nature provides an elegant answer through wave propagation~\cite{sorteberg2019approximating}. The wave equation, governing phenomena from acoustic waves to electromagnetic fields, describes how information propagates through space with remarkable properties: different frequency components evolve independently according to their natural frequencies, energy is conserved during propagation, and the entire wave field can be reconstructed from its spectral components. These characteristics align remarkably well with the requirements of SR—frequency-dependent processing naturally separates high-frequency details from low-frequency structures, while energy conservation ensures stable feature propagation.

Inspired by this physical insight, we propose WaveSR, a physics-inspired wave evolution attention that integrates wave equation dynamics into lightweight SR networks. Our key innovation lies in formulating attention computation as wave evolution in the frequency domain. 
This approach offers two compelling advantages:
First, by leveraging the inherent frequency separation property of wave propagation, it achieves targeted feature enhancement without introducing additional complex modules. Second, the proposed wave evolution paradigm endows the network with intrinsic physical interpretability, breaking the ``black-box" limitation of conventional deep learning-based SR models and rendering feature modulation during image reconstruction more controllable and predictable. As shown in Fig.~\ref{fig:intro}, to the best of our knowledge, WaveSR is among the first studies to investigate the use of partial differential equation solutions for efficient image super-resolution. The results demonstrate the potential of physics-inspired feature modeling for lightweight SR. Extensive experiments show that WaveSR delivers competitive reconstruction quality while achieving a favorable trade-off between model complexity and reconstruction performance.

Our contributions can be summarized as follows:

1. We propose WaveSR, a lightweight image super-resolution method that introduces an evolution mechanism inspired by the wave equation to model frequency-domain feature interactions and facilitate high-frequency detail reconstruction.

2. We design the Wave Evolution Attention (WEA) module, which models feature interactions through the analytical solution of the 2D wave equation in the Fast Fourier Transform (FFT) domain, allowing independent evolution of different frequency components with learnable wave speed \(c\) and damping coefficient \(\alpha\).

3. We propose a Frequency-Adaptive Feed-Forward Network (FA-FFN) that achieves adaptive frequency feature selection through learnable spectral modulation, effectively enhancing feature representation capability.

\section{Related Works}

\subsection{Deep Networks for Image Super-Resolution}
Deep neural networks have emerged as the dominant paradigm for Image SR due to their exceptional representation learning capabilities. Since SRCNN~\cite{srcnn} first introduced convolutional neural networks to SR with a simple three-layer architecture, a plethora of CNN-based methods have pushed the state-of-the-art through increasingly sophisticated designs. Notable advancements include residual connections~\cite{lim2017enhanced, liu2020residual,ledig2017photo, zhang2019ranksrgan} to facilitate training of deeper networks, and U-shaped architectures~\cite{cheng2019encoder,liu2018multi,mao2016image,kong2023efficient} that effectively aggregate multi-scale features through encoder-decoder pathways. Recognizing the limitations of CNNs in modeling long-range dependencies, attention mechanisms have been extensively explored for SR. Channel attention~\cite{zhang2018image} adaptively recalibrates feature responses, while spatial attention complements this by focusing on informative regions. Methods like HASN~\cite{cao2025hasn} and DAT~\cite{chen2023dual} combine both spatial and channel attention to capture more comprehensive feature interactions. More recently, Transformer-based architectures have revolutionized SR by leveraging self-attention's ability to model global contexts. SwinIR~\cite{liang2021swinir} adapts the Swin Transformer~\cite{liu2021swin} framework to SR through window-based attention, dividing images into 8×8 windows with shifted window partitioning to enable cross-window information exchange. HAT~\cite{chen2023hat} enhances the interaction between neighboring window features through an overlapping cross-attention module. Recently, the linear complexity of Mamba has attracted widespread attention and has been progressively applied to image restoration tasks~\cite{lei2024dvmsr,guo2024mambair,guo2025mambairv2}.

\subsection{Lightweight Super-Resolution Methods}
The growing demand for deploying SR models on resource-constrained platforms has catalyzed research into lightweight SR architectures~\cite{li2022blueprint,shi2016real,zheng2024smfanet,sun2023spatially}. Early approaches such as ESPCN~\cite{shi2016real} and FSRCNN~\cite{fsrcnn} adopted post-upsampling strategies to reduce computational overhead by processing features in the low-resolution space.
Information distillation networks~\cite{IDN2018,hui2019lightweight,liu2020residual} partition input features into multiple streams, with each stream progressively refined before fusion for rich representation learning. In the Transformer era, lightweight SR research has focused on sparse attention mechanisms to mitigate the quadratic complexity of self-attention. Window-based attention~\cite{liang2021swinir,chen2023hat} restricts self-attention computation to local windows, while group-wise attention~\cite{zhang2022efficient} partitions channels into groups for parallel processing. CATANet~\cite{liu2025catanet} and SPIN~\cite{zhang2023lightweight} aggregate long-range content-similar tokens to provide dense attention. MAT~\cite{xie2025mat} proposes a multi-range attention Transformer that incorporates dilated convolution principles into self-attention, enabling the capture of multi-scale dependencies across local, regional, and sparse global ranges. CFIN~\cite{cfin} achieves dynamic contextual feature extraction by fusing CNN and Transformer architectures.
Moreover, the advantages of frequency-domain representations are increasingly being explored for image super-resolution. FourierSR~\cite{li2026fouriersr} performs channel-token mixing and applies broadcast local filters separately to the real and imaginary Fourier components, yielding an efficient global dynamic convolution based on the convolution theorem. DMNet~\cite{dmnet} combines wavelet and Fourier information to balance structural reconstruction and high-frequency detail restoration. HCSA~\cite{hcsa} employs Haar wavelet encoding to replace standard downsampling operations, thereby avoiding the loss of important information. In this work, We also use the frequency-domain to implement a structured, frequency-dependent wave evolution operator for content-conditioned feature interaction.

\subsection{Physics-inspired Neural Networks}
Beyond conventional architectural innovations, an emerging paradigm integrates physical principles into neural network design, offering both improved performance and interpretability. Physics-informed neural networks~\cite{raissi2019physics} embed governing equations as soft constraints during training, while neural operators~\cite{kovachki2023neural} learn mappings between function spaces by respecting physical symmetries. 
vHeat~\cite{wang2025building} conceptualizes image patches as heat sources and models their correlations through adaptive thermal diffusion, enabling efficient global context modeling. Wave propagation has been shown to offer great potential for modeling signal evolution with balanced frequency characteristics~\cite{sorteberg2019approximating, rasht2022physics, shu2026waveformer}. Among these methods, WaveFormer~\cite{shu2026waveformer} evolves a single semantic field globally through a wave-equation-derived propagation operator. Such models preserve both global semantics and local textures, providing a physics-inspired alternative to purely data-driven architectures.
These physics-inspired approaches leverage the mathematical rigor of physical laws while maintaining the flexibility of deep learning.
In low-level vision, recent works have drawn inspiration from classical physical models. For instance, diffusion models~\cite{mei2025power, duan2025dit4sr, li2025difiisr, yue2025arbitrary} simulate reverse-time stochastic processes for image generation. SSIU~\cite{ni2025structural} introduces a structural similarity-inspired unfolding network that decomposes the MAP super-resolution problem via half-quadratic splitting into iterative feature-space updates. Following this line of physically motivated modeling, we exploit wave-equation-based vision modeling to reformulate global semantic propagation as local, content-conditioned feature interaction.

\begin{figure*}[!tbp]
\centering
\includegraphics[width=6.5in]{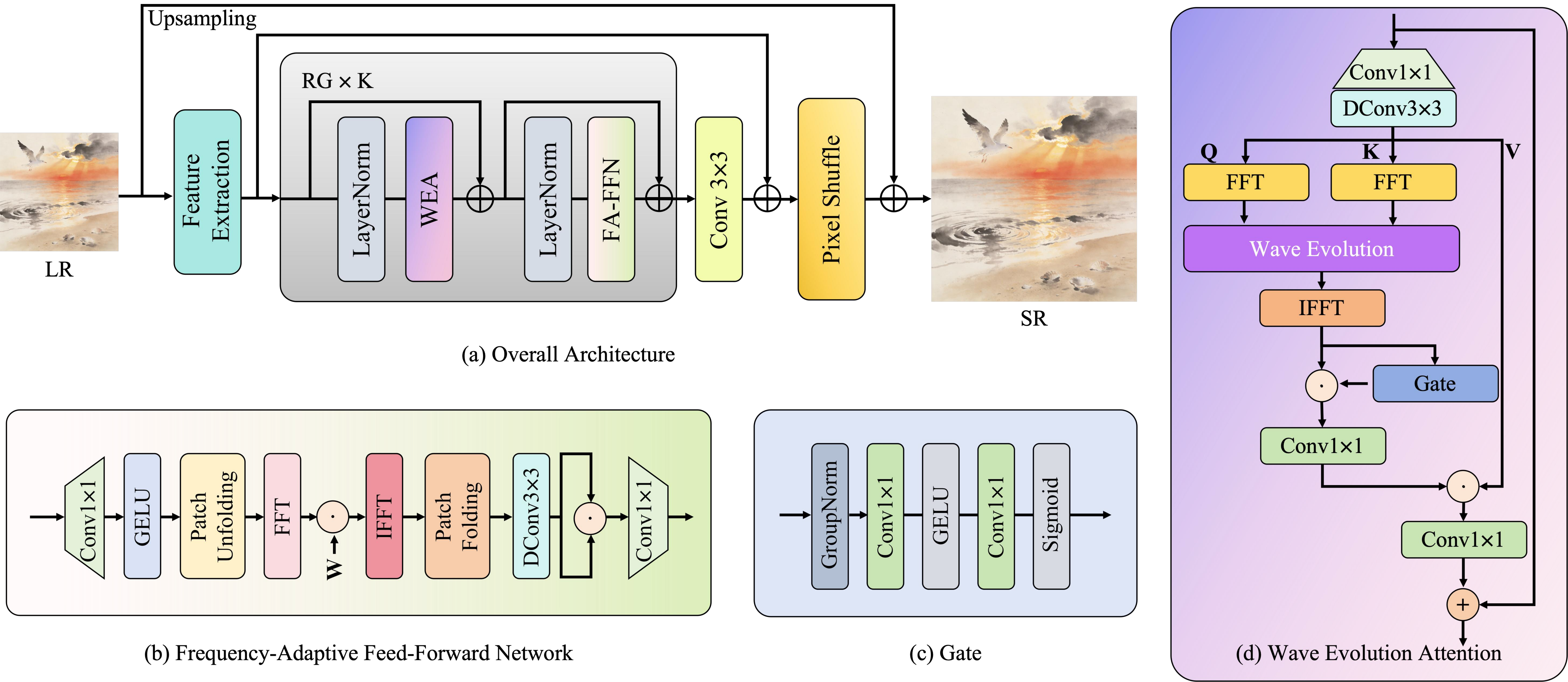}
\caption{The overall architecture of WaveSR and the structure of Wave Evolution Attention and Frequency-Adaptive Feed-Forward Network.}
\label{fig:hifiwave}
\end{figure*}

\section{Preliminaries}
The two-dimensional damped wave equation describes the evolution of a scalar field $u(x,y,t)$ as
\begin{equation}
\frac{\partial^2 u}{\partial t^2}
+\alpha\frac{\partial u}{\partial t}
=
v^2\left(
\frac{\partial^2u}{\partial x^2}
+
\frac{\partial^2u}{\partial y^2}
\right),
\label{eq:damped_wave}
\end{equation}
where $t$ denotes an evolution variable, $v>0$ is the wave speed, and $\alpha\geq0$ is the damping coefficient. For feature modeling, this equation can be applied independently to each channel of a latent feature field.

Applying the two-dimensional spatial Fourier transform converts Eq.~\eqref{eq:damped_wave} into an ordinary differential equation for each spatial-frequency mode:
\begin{equation}
\frac{\partial^2\hat{u}}{\partial t^2}
+\alpha\frac{\partial\hat{u}}{\partial t}
+v^2k^2\hat{u}
=0,
\label{eq:frequency_ode}
\end{equation}
where $k=\sqrt{k_x^2+k_y^2}$ is the radial wavenumber. For normalized spatial frequencies $f_x,f_y$, we have $k_x=2\pi f_x$ and $k_y=2\pi f_y$. Given the initial conditions
$\hat{\phi}=\mathcal{F}[u(\cdot,\cdot,0)]$ and
$\hat{\psi}=\mathcal{F}[u_t(\cdot,\cdot,0)]$, an underdamped mode satisfying $v^2k^2>(\alpha/2)^2$ evolves as
\begin{equation}
\hat{u}(k_x,k_y,t)
=
e^{-\frac{\alpha}{2}t}
\left[
\hat{\phi}\cos(\omega_dt)
+
\frac{\hat{\psi}+\frac{\alpha}{2}\hat{\phi}}
{\omega_d}
\sin(\omega_dt)
\right],
\label{eq:frequency_solution}
\end{equation}
where
$\omega_d=
\sqrt{v^2k^2-\left(\frac{\alpha}{2}\right)^2}$
is the damped oscillation frequency. Critically damped and overdamped modes, including the DC mode when $\alpha>0$, follow the corresponding limiting and hyperbolic forms.

The spatial Fourier transform allows different frequency modes to evolve independently, while $\omega_d(k)$ introduces a frequency-dependent oscillatory response. For underdamped modes, $e^{-\alpha t/2}$ provides a frequency-independent decay envelope and therefore does not impose stronger exponential attenuation on higher spatial frequencies. The cosine-sine evolution structure in Eq.~\eqref{eq:frequency_solution} provides the theoretical basis for constructing the frequency-domain feature interaction in WaveSR.
\begin{figure}[!tbp]
\centering
\includegraphics[width=3in]{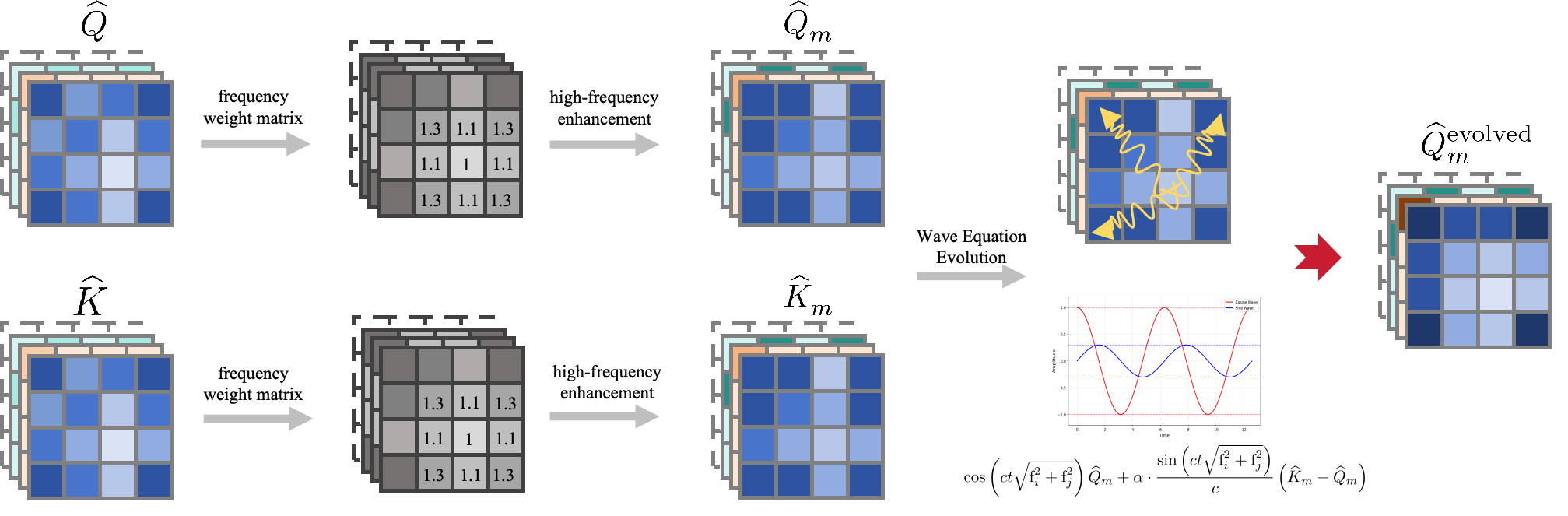}
\caption{The implementation process of wave evolution.}
\label{fig:WE}
\end{figure}
\section{The Proposed WaveSR}
The proposed WaveSR consists of a shallow feature extraction layer, several stacked residual groups, and a pixel shuffle layer, as illustrated in Fig.~\ref{fig:hifiwave}(a). The shallow feature extraction is performed by a \(3 \times 3\) convolutional layer, which maps the image from the original input space to a high-dimensional feature space. Let \(\mathbf{I}_{LR}\) and \(\mathbf{I}_{SR}\) represent the LR input image and the SR image, respectively. 
We first obtain the shallow feature \(\mathbf{X}_0\) by
\begin{equation}
\mathbf{X}_0 = \mathbf{F}_S(\mathbf{I}_{LR}),
\end{equation}
where \(\mathbf{F}_S\) denotes the shallow feature extraction module. Then, \(\mathbf{X}_0\) is fed into multiple residual connection layers to extract deep features, which can be expressed as:

\begin{equation}
\mathbf{X}_D = \mathbf{F}_{RG}(\mathbf{X}_0),
\end{equation}
where \(\mathbf{F}_{RG}\) represents the residual groups. Subsequently, a \(3 \times 3\) convolutional layer is applied to aggregate the deep features:

\begin{equation}
\mathbf{X}_A = \mathbf{F}_{\text{conv}}(\mathbf{X}_D).
\end{equation}
A long skip connection is then used to combine the shallow feature \(\mathbf{X}_0\) with the aggregated feature \(\mathbf{X}_A\):

\begin{equation}
\mathbf{X}_C = \mathbf{X}_0 + \mathbf{X}_A.
\end{equation}
The high-resolution image is reconstructed via a pixel shuffle layer~\cite{shi2016real}. Finally, the original input \(\mathbf{I}_{LR}\) is upsampled to the same spatial size as the output of the pixel shuffle layer, and another long skip connection is employed to aggregate these features, producing the final SR image.

\subsection{Wave Evolution Attention}
Inspired by the solution to wave equations, we propose an attention module termed Wave Evolution Attention (WEA). An overview of the proposed module is shown in Fig.~\ref{fig:hifiwave}(d).
Given an input feature map $\mathbf{X} \in \mathbb{R}^{B \times C \times H \times W}$, we first generate query, key, and value features through lightweight convolutions:

\begin{equation}
\mathbf{Q}, \mathbf{K}, \mathbf{V} = \text{Chunk}(\text{DWConv}_{3\times3}(\text{Conv}_{1\times1}(\mathbf{X})), 3),
\end{equation}
where each of $\mathbf{Q}, \mathbf{K}, \mathbf{V}$ maintains $C$ channels. To capture local frequency characteristics while maintaining computational efficiency, we partition the feature maps into non-overlapping patches of size $p \times p$:

\begin{equation}
\mathbf{Q}_p, \mathbf{K}_p, \mathbf{V}_p \in \mathbb{R}^{N \times C \times p \times p}, \quad N = \frac{H}{p} \cdot \frac{W}{p},
\end{equation}
we transform the query and key patches to the frequency domain using the 2D Fast Fourier Transform (FFT) with centering to align low frequencies at the center:

\begin{equation}
\hat{\mathbf{Q}} = \mathcal{F}_{\text{shift}}(\mathcal{F}_{\text{fft2}}(\mathbf{Q}_p)), \quad \hat{\mathbf{K}} = \mathcal{F}_{\text{shift}}(\mathcal{F}_{\text{fft2}}(\mathbf{K}_p)),
\end{equation}
where $\mathcal{F}_{\text{fft2}}$ denotes the 2D FFT and $\mathcal{F}_{\text{shift}}$ centers the frequency spectrum. This centering operation ensures that low-frequency components (corresponding to structural information) are positioned at the center, while high-frequency components (carrying texture details) are distributed toward the edges, facilitating frequency-aware processing. Subsequently, high-frequency and low-frequency components evolve independently via the Wave Evolution (WE) process, which can be expressed as:

\begin{equation}
\hat{\mathbf{Q}}' = \text{WE}(\hat{\mathbf{Q}}, \hat{\mathbf{K}}),
\end{equation}
where \(\text{WE}(\cdot)\) represents the wave evolution process. To transform the features back to the spatial domain, we apply an inverse Fast Fourier Transform (IFFT) followed by inverse shift operations.

\begin{equation}
\mathbf{Q}' = \mathcal{F}_{\text{fft2}}^{-1}\left(\mathcal{F}_{\text{shift}}^{-1}(\hat{\mathbf{Q}}')\right),
\end{equation}
where $\mathcal{F}_{\text{fft2}}^{-1}$ is the inverse 2D FFT and $\mathcal{F}_{\text{shift}}^{-1}$ is the inverse shift operation. We use a lightweight frequency-guided gating mechanism:

\begin{equation}
\mathbf{G} = \sigma(\text{Conv}_{1\times1}(\text{GELU}(\text{Conv}_{1\times1}(\text{GN}(\mathbf{Q}'))))),
\end{equation}
\begin{equation}
\mathbf{Q}_g = \mathbf{Q}' \odot \mathbf{G},
\end{equation}
where $\text{GN}$ denotes Group Normalization and $\sigma$ is the Sigmoid activation. $\odot$ is hadamard product. This gating mechanism uses $1\times1$ convolution to reduce parameter overhead while enabling adaptive feature selection.
To further emphasize high-frequency details, we apply a $1\times1$ convolution, and the final output $\mathbf{X_{att}}$ is obtained through element-wise modulation of the value features, followed by a residual connection:

\begin{equation}
\mathbf{X_{att}} = \text{Conv}_{1\times1}(\mathbf{V} \odot \text{Conv}_{1\times1}(\mathbf{Q}_g)) + \mathbf{X}
\end{equation}

\noindent\textbf{Wave Evolution.}
The detailed implementation of Wave Evolution is illustrated in Fig.~\ref{fig:WE}. Recognizing that super-resolution primarily benefits from high-frequency detail recovery, we first introduce a learnable radial mask to adaptively reweight the query and key spectra, thereby emphasizing detail-rich components:

\begin{equation}
\hat{\mathbf{Q}}_m = \hat{\mathbf{Q}} \odot \mathbf{M}, \quad \hat{\mathbf{K}}_m = \hat{\mathbf{K}} \odot \mathbf{M},
\end{equation}
where $\mathbf{M} \in \mathbb{R}^{p \times p}$ assigns weights according to the spectral distance from the center:

\begin{equation}
M_{u,v} = 1 + (\gamma - 1) \cdot \frac{\sqrt{(u - u_0)^2 + (v - v_0)^2}}{\max(\text{dist})},
\end{equation}
with $u_0 = v_0 = p/2$ being the center index, $\gamma = 1.2$ the high-frequency enhancement coefficient, and $\max(\text{dist})$ normalizing the distance to $[0,1]$. This formulation progressively increases weights from low to high frequencies, effectively prioritizing the detail-rich frequency bands critical for super-resolution.

Inspired by the frequency-domain solution of the two-dimensional wave equation, we formulate cross-feature interaction as a wave propagation process. The analytical solution is given by:

\begin{equation}
\hat{\mathbf{U}}(t) = \cos(c_w \boldsymbol{\omega} t) \odot \hat{\mathbf{U}}_0 + \frac{\sin(c_w \boldsymbol{\omega} t)}{c_w |\boldsymbol{\omega}|} \odot \frac{\partial \hat{\mathbf{U}}}{\partial t}\bigg|_0,
\end{equation}
where $\boldsymbol{\omega}$ denotes the angular frequency matrix. In practice, we omit the $1/|\boldsymbol{\omega}|$ term to avoid numerical singularities near zero frequency and to simplify complex-valued computations, while preserving the essential modulation behavior of wave propagation. To bridge this physical process with feature evolution in our attention module, we treat the masked query spectrum $\hat{\mathbf{Q}}_m$ as the initial displacement $\hat{\mathbf{U}}_0$, and the residual $\hat{\mathbf{K}}_m - \hat{\mathbf{Q}}_m$ as the initial velocity $\frac{\partial \hat{\mathbf{U}}}{\partial t}\big|_0$ that drives the evolution. This conceptual mapping is motivated by two considerations. First, from the perspective of attention mechanisms, the residual $\hat{\mathbf{K}}_m - \hat{\mathbf{Q}}_m$ directly reflects the feature discrepancy between the query and key projections of the same input, serving as a natural driving force for feature refinement. Second, from a signal processing viewpoint, since $\hat{\mathbf{Q}}_m$ and $\hat{\mathbf{K}}_m$ largely share low-frequency components, their difference predominantly resides in the high-frequency regime. This makes the residual inherently rich in high-frequency details, which precisely corresponds to the perturbation energy carried by the velocity term in the wave equation.
For each spectral component, the angular frequency is defined as:

\begin{equation}
\omega_{u,v} = 2\pi \sqrt{f_u^2 + f_v^2},
\end{equation}
where $f_u, f_v \in [-0.5, 0.5]$ are the centered normalized spatial frequencies. Substituting the above definitions into the wave equation solution and applying the simplification yields our wave evolution rule:

\begin{equation}
\hat{\mathbf{Q}}' = \cos(c t \boldsymbol{\omega}) \odot \hat{\mathbf{Q}}_m + \alpha \cdot \frac{\sin(c t \boldsymbol{\omega})}{c} \odot (\hat{\mathbf{K}}_m - \hat{\mathbf{Q}}_m).
\end{equation}

In the physical wave equation, $c$ and $t$ correspond to wave speed and propagation time, respectively. In our latent feature space, however, these quantities no longer carry physical units. We therefore treat $c$ and $t$ as learnable parameters that control the effective spectral modulation scale, while $\alpha$ is a learnable coefficient that adjusts the contribution of the velocity term.
All operations are performed in the complex domain via 2D FFT, with element-wise multiplications and trigonometric computations applied separately to the real and imaginary parts to ensure numerical stability.

\subsection{Frequency-Adaptive Feed-Forward Network}

The Feed-Forward Network (FFN) plays a crucial role in enhancing features after attention computation, directly influencing the quality of reconstructed images. While standard FFNs~\cite{dosovitskiy2020image, vaswani2017attention, restormer} process all frequency components uniformly, not all frequency information contributes equally to latent clear image restoration. Low-frequency components provide structural stability, while high-frequency components carry texture details. However, both may contain redundant or detrimental information. Therefore, it is essential to develop a frequency-adaptive FFN that can selectively preserve relevant frequency bands while suppressing irrelevant ones. We propose a Frequency-Adaptive Feed-Forward Network (FA-FFN) that introduces learnable frequency modulation through a differentiable spectral transformation. Unlike conventional FFNs that operate purely in the spatial domain, our method explicitly models frequency importance via learnable weights in the Fourier domain, enabling content-aware frequency selection. The proposed FA-FFN can be formulated as:
\begin{equation}
\begin{aligned}
&\mathbf{X} =  \text{Conv}_{1\times1}(\mathbf{X}_{\text{att}}) \in \mathbb{R}^{B \times 2C' \times H \times W}, \\
&\mathbf{X}_p = \mathcal{P}(\mathbf{X}) \in \mathbb{R}^{B \times 2C' \times \frac{H}{p} \times \frac{W}{p} \times p \times p}, \\
&\hat{\mathbf{X}}_p = \mathcal{F}_{\text{rfft}}(\mathbf{X}_p) \in \mathbb{C}^{B \times 2C' \times \frac{H}{p} \times \frac{W}{p} \times p \times (\frac{p}{2}+1)}, \\
&\hat{\mathbf{X}}_p' = \hat{\mathbf{X}}_p \odot \mathbf{W}_{\text{fft}}, \\
&\mathbf{X}_p' = \mathcal{F}_{\text{irfft}}(\hat{\mathbf{X}}_p', s=(p,p)), \\
&\mathbf{X}' = \mathcal{P}^{-1}(\mathbf{X}_p') \in \mathbb{R}^{B \times 2C' \times H \times W}, \\
&\mathbf{X}_{\text{dw}} = \text{DWConv}_{3\times3}(\mathbf{X}') \in \mathbb{R}^{B \times 2C' \times H \times W}, \\
&\mathbf{X}_{\text{out}} =  \text{Conv}_{1\times1}(\mathcal{G}(\mathbf{X}_{\text{dw}})),
\end{aligned}
\end{equation}
where $\mathcal{P}(\cdot)$ and $\mathcal{P}^{-1}(\cdot)$ denote patch unfolding and folding operations with patch size $p$. $\mathcal{F}_{\text{rfft}}(\cdot)$ and $\mathcal{F}_{\text{irfft}}(\cdot)$ represent the Real Fast Fourier Transform (RFFT) and its inverse. For each $p \times p$ patch, 2D RFFT is applied, yielding frequency representations with the last dimension reduced to $p/2+1$ due to conjugate symmetry. $\mathbf{W}_{\text{fft}} \in \mathbb{C}^{2C' \times p \times (\frac{p}{2}+1)}$ is a learnable complex weight tensor that modulates each frequency component. $\odot$ denotes element-wise complex multiplication. $\text{DWConv}_{3\times3}$ is depthwise convolution for spatial context aggregation. $\mathcal{G}(\cdot)$ denotes the Gated Linear Unit (GLU), which splits the input into two equal parts along the channel dimension and multiplies them element-wise. $\mathbf{X}_{\text{att}}$ is the input feature from the WEA module. $C' = \gamma C$ with expansion factor $\gamma$. The learnable frequency weights $\mathbf{W}_{\text{fft}}$ play a role analogous to the quantization matrix in JPEG compression, but with learnable weights that can either preserve or enhance frequency components. This frequency-adaptive mechanism ensures that the FFN emphasizes informative spectral components while suppressing less useful ones, leading to more effective feature refinement. The detailed network architecture of the proposed FA-FFN is illustrated in Fig.~\ref{fig:hifiwave}(b).

\subsection{Loss Function}
Our loss function comprises two components: a spatial-domain pixel loss and a frequency-domain loss, defined as:

\begin{equation}
\mathcal{L}_{total} = \lambda_{1}\mathcal{L}_{{pixel}} + \lambda_{freq} \mathcal{L}_{{freq}},
\label{eq:loss}
\end{equation}
where \(\lambda_{1}\) is the weight of $\mathcal{L}_{{pixel}}$ and \(\lambda_{freq}\) is the weight of $\mathcal{L}_{{freq}}$.
\subsubsection{Pixel Loss}
The pixel loss is implemented using $\mathcal{L}_{1}$ loss:

\begin{equation}
\mathcal{L}_{{pixel}} = \|I_{SR} - I_{HR}\|_1,
\end{equation}
where \(I_{SR}\) and \(I_{HR}\) denote the predicted and ground-truth images, respectively.

\subsubsection{Frequency-Domain Loss}

The frequency-domain loss complements the spatial-domain pixel loss by explicitly constraining the spectral discrepancy between the reconstructed image and its ground truth. Given the predicted image $I_{SR}$ and the ground-truth image $I_{HR}$, we first transform them into the frequency domain using the orthonormally normalized real-valued two-dimensional Fourier transform:

\begin{equation}\widehat{I}_{SR} = \mathcal{F}_{\mathrm{rfft}}(I_{SR}), \qquad \widehat{I}_{HR} = \mathcal{F}_{\mathrm{rfft}}(I_{HR}).\end{equation}

The magnitude of their complex frequency-domain residual is calculated as
\begin{equation}\mathbf{D}_{b,c,h,w} = \left|\widehat{I}_{SR}^{b,c}(h,w) - \widehat{I}_{HR}^{b,c}(h,w)\right|,\end{equation}
where $b$ and $c$ denote the batch and channel indices, respectively, while $h$ and $w$ index the frequency-domain coefficients. This formulation calculates the magnitude of the complex spectral residual. To construct the frequency masks, we define a coordinate system on the RFFT output grid of size $H \times (\lfloor W/2 \rfloor + 1)$:
\begin{equation}u_h = h - \left\lfloor\frac{H}{2}\right\rfloor, \qquad v_w = w,\end{equation}
where $h \in [0, H-1]$ and $w \in [0, \lfloor W/2 \rfloor]$. Note that for the width dimension, the RFFT only retains non-negative frequencies, so $v_w$ directly uses the frequency indices without centering. The normalized radial distance is subsequently calculated as
\begin{equation}r_{h,w} = \frac{\sqrt{u_h^2 + v_w^2}}{\sqrt{\left\lfloor H/2\right\rfloor^2 + \left\lfloor W/2\right\rfloor^2}}.\end{equation}

Based on this normalized radial distance, we construct three disjoint masks directly on the RFFT grid:
\begin{equation}\begin{aligned}
\mathbf{M}_{L}(h,w) &= \mathbb{I}\left(r_{h,w} \leq \tau_L \right), \\
\mathbf{M}_{M}(h,w) &= \mathbb{I}\left(\tau_L < r_{h,w} \leq \tau_H\right), \\
\mathbf{M}_{H}(h,w) &= \mathbb{I}\left(r_{h,w} > \tau_H\right),
\end{aligned}\end{equation}
where $\mathbb{I}(\cdot)$ denotes the indicator function, $0 < \tau_L < \tau_H < 1$. 
The thresholds $\tau_L$ and $\tau_H$ divide the normalized radial range into three approximately equal-width intervals. 
This fixed partition remains resolution-adaptive because the normalized radial distance is defined relative to the image dimensions, and it does not introduce additional learnable parameters.
For each frequency region, the masked spectral residual is defined as
\begin{equation}\mathbf{D}_{i} = \mathbf{D} \odot \mathbf{M}_{i},\qquad i \in \{L, M, H\}.\end{equation}

Following the implementation, for each frequency region, the L1 error is summed over the frequency coefficients and then averaged over the batch and channel dimensions:

\begin{equation}\mathcal{L}_{i} = \frac{1}{BC}\sum_{b=1}^{B}\sum_{c=1}^{C}\sum_{h,w} \left(\mathbf{D} \odot \mathbf{M}_{i}\right)_{b,c,h,w}.\end{equation}

The final frequency-domain loss is defined as
\begin{equation}
\mathcal{L}_{freq} =w_L \mathcal{L}_{L} + w_M \mathcal{L}_{M} + w_H \mathcal{L}_{H},
\end{equation}
where $w_L$, $w_M$, and $w_H$ control the supervision strengths of the low-, middle-, and high-frequency regions, respectively.

\section{Experiments}

\begin{table*}[!htpb]
\small
\centering
\begin{center}
\caption{Benchmark results with bicubic degradation. We show average PSNR (dB) \& SSIM values for scaling factor $\times4$. We highlight the top-2 least params, FLOPs, and performance on each datasets with \textbf{bold} (best) and \textit{italics} (second), respectively. + denotes the results are generated with self-ensemble.}
\scalebox{0.85}{

\begin{tabular}{l |c|c|c|c|c|c|c|c|c|c|c|cc}
\hline
\multirow{2}{*}{Method Name} & \multirow{2}{*}{\#Params} & \multirow{2}{*}{FLOPs} & \multicolumn{2}{c}{Set5} & \multicolumn{2}{|c}{Set14} & \multicolumn{2}{|c}{BSD100} & \multicolumn{2}{|c}{Urban100} & \multicolumn{2}{|c}{Manga109}\\
\cline{4-13}
 & & & PSNR & SSIM & PSNR & SSIM & PSNR & SSIM & PSNR & SSIM & PSNR & SSIM \\
\hline
\hline
Bicubic & - & - & 28.42 & 0.8104 & 26.00 & 0.7027 & 25.96 & 0.6675 & 23.14 & 0.6577 & 24.89 & 0.7866 \\

VDSR~\cite{kim2016accurate} & 665K & 612.6G & 31.35 & 0.8830 & 28.02 & 0.7680 & 27.29 & 0.7260 & 25.18 & 0.7540 & 28.83 & 0.8870\\

DRCN~\cite{kim2016deeply} & 1774K & 9788.7G & 31.53 & 0.8854 & 28.02 & 0.7670 & 27.23 & 0.7233 & 25.18 & 0.7524 & 28.93 & 0.8854\\

LapSRN~\cite{lai2017deep} & 813K & 149.4G & 31.54 & 0.8850 & 28.19 & 0.7720 & 27.32 & 0.7280 & 25.21 & 0.7560 & 29.09 & 0.8845\\

MemNet~\cite{tai2017memnet} & 677K & 623.9G & 31.74 & 0.8893 & 28.26 & 0.7723 & 27.40 & 0.7281 & 25.50 & 0.7630 & 29.42 & 0.8942\\

CARN~\cite{CARN2018} & 1592K & 90.9G & 32.13 & 0.8937 & 28.60 & 0.7806 & 27.58 & 0.7349 & 26.07 & 0.7837 & 30.47 & 0.9084\\

IDN~\cite{IDN2018} & 553K & 32.3G & 31.82 & 0.8903 & 28.25 & 0.7730 & 27.41 & 0.7297 & 25.41 & 0.7632 & 29.41 & 0.8942\\

SRFBN-S~\cite{SRFBN-S2019} & 483K & 852.9G & 31.98 & 0.8923 & 28.45 & 0.7779 & 27.44 & 0.7313 & 25.71 & 0.7719 & 29.91 & 0.9008\\

IMDN~\cite{hui2019lightweight} & 715K & 40.9G & 32.21 & 0.8948 & 28.58 & 0.7811 & 27.56 & 0.7353 & 26.04 & 0.7838 & 30.45 & 0.9075 \\


LatticeNet~\cite{latticenet2020} & 777K & 43.6G & 32.18 & 0.8943 & 28.61 & 0.7812 & 27.57 & 0.7355 & 26.14 & 0.7844  & - & - \\



RFDN~\cite{liu2020residual} & 550K & 31.6G & 32.24& 0.8952 & 28.61& 0.7819&  27.57& 0.7360 & 26.11& 0.7858 &  30.58 & 0.9089\\

SMSR~\cite{SMSR2021} & 1006K & 41.6G & 32.12 & 0.8932 & 28.55 & 0.7808 & 27.55 & 0.7351 & 26.11 & 0.7868 & 30.54 & 0.9085 \\

RLFN~\cite{kong2022residual} & 543K  & 33.9G &  32.24 & 0.8952 &  28.62 & 0.7813 &  27.60 & 0.7364  &  \textit{26.17} &  \textit{0.7877} & – & – \\

ShuffleMixer~\cite{sun2022shufflemixer}& \textit{411K}  & 28G & 32.21 & 0.8953 &   28.66&0.7827& 27.61&0.7366 &26.08&0.7835& 30.65&0.9093\\


HPUN-M~\cite{sun2023hybrid} & 511K & 27.7G &  \textit{32.24} & 0.8950 & 28.66 & 0.7828 & 27.60 & 0.7371 & 26.12 & \textbf{0.7878} & 30.55 &  \textit{0.9089} \\

DVMSR~\cite{lei2024dvmsr}& 424K & \textbf{20.2G} & 32.19 & 0.8955 & 28.61 & 0.7823 & 27.58 & 0.7379 & 26.03 & 0.7838 &  30.48 & 0.9084 \\

HASN~\cite{cao2025hasn}  & 435K  & 26.6G & 32.23 &  \textit{0.8960} &  \textit{28.66} &  0.7830 &   27.62 &  {0.7387}  & 26.13 & 0.7869 &  30.50 & 0.9077\\

SRConvNet~\cite{srconvnet} & \textbf{382K}  & \textit{22G} & 32.18 &  0.8951 &   28.61 &  0.7818 &  27.57  &  {0.7359}  & 26.06 & 0.7845 &  30.35 & 0.9075\\

WaveSR & 414K &22.4G &  32.23 & 0.8959 & 28.64  &  \textit{0.7830}& \textit{27.62}  & \textit{0.7388} & 26.11 & 0.7847 &  \textit{30.66}  & 0.9088 \\

WaveSR+ & 414K &22.4G &  \textbf{32.27} & \textbf{0.8967} & \textbf{28.69}  &  \textbf{0.7842}& \textbf{27.65}  & \textbf{0.7396} & \textbf{26.19} & 0.7867 &  \textbf{30.82}  & \textbf{0.9108} \\

\hline
\end{tabular}}
\label{tab:results}
\end{center}
\end{table*}

\subsection{Experimental Settings}
\subsubsection{Datasets and Metrics}
Following common practices in image super-resolution~\cite{long2025progressive,cao2025hasn}, we train our models on the DF2K dataset, which combines DIV2K~\cite{div2k} and Flickr2K~\cite{div2k} to provide a diverse collection of high-quality images. During training, we synthesize LR inputs by bicubic downsampling with $\times 4$ scale factors. For data augmentation, we employ random horizontal flips and rotations. For evaluation, we use five standard benchmark datasets: Set5~\cite{Set5}, Set14~\cite{Set14}, BSD100~\cite{B100}, Urban100~\cite{Urban100}, and Manga109~\cite{Manga109}. 
We evaluate reconstruction quality using three widely adopted metrics: PSNR, SSIM~\cite{ssim} and LPIPS~\cite{lpips}. Following conventional practice~\cite{liang2021swinir,chen2023hat}, we compute these metrics on the luminance channel (Y) of the YCbCr color space.  Higher PSNR and SSIM values indicate better reconstruction quality. For LPIPS, lower values indicate higher perceptual similarity to the ground truth. We primarily employ LPIPS in our comparisons with real-world SR methods, where perceptual quality is of greater concern than pixel-wise fidelity.
\subsubsection{Implementation Details}
\noindent\textbf{Training details.}
The entire training process consists of two stages. In the first stage, we train only the FA-FFN without the WEA module, employing the $\mathcal{L}_{{pixel}}$. This stage is conducted for 400k iterations, and the batch size is 4. In the second stage, we initialize the network with the model trained in the first stage and adopt our proposed loss function $\mathcal{L}_{total}$ (as defined in Eq.~(\ref{eq:loss})), where the weight of the frequency loss term $\lambda$ is set to 1. This stage runs for 500k iterations, and the batch size is 64.
Both stages use the Adam optimizer with parameters \(\beta_1 = 0.9\) and \(\beta_2 = 0.99\). The initial learning rate is set to $2 \times10^{-4}$ and is halved at milestones of 250k, 400k, 450k, and 475k iterations. All experiments are conducted using the PyTorch framework on an NVIDIA GeForce RTX 3090 GPU. FLOPs is measured when the input is $3\times 256\times 256$.
\\
\noindent\textbf{Architecture details.} WaveSR is composed of 6 residual groups, each of which contains a WEA module and a FA-FFN module. The feature dimension is set to 56, and the patch size is $64\times 64$. In the wave evolution process, both the damping coefficient \(\alpha\) and the wave speed \(c\) are designed as learnable parameters. Specifically, the damping coefficient \(\alpha\) is initialized to 0.1 to control the overall intensity of the evolution process, while the wave speed \(c\) is initialized to 1.0 to regulate the propagation range. The time step \(t\) is initialized to 1.0. The thresholds $\tau_L$ and $\tau_H$ are set to 0.33 and 0.66, respectively. For more detailed parameter configurations, please refer to our open-source code: \url{https://github.com/nathan66666/WaveSR.git}.

\subsection{Comparisons with the State-of-the-Art}
\subsubsection{Comparison with Lightweight SR Methods}
\begin{figure}[!htbp]
\centering
\includegraphics[width=3.3in]{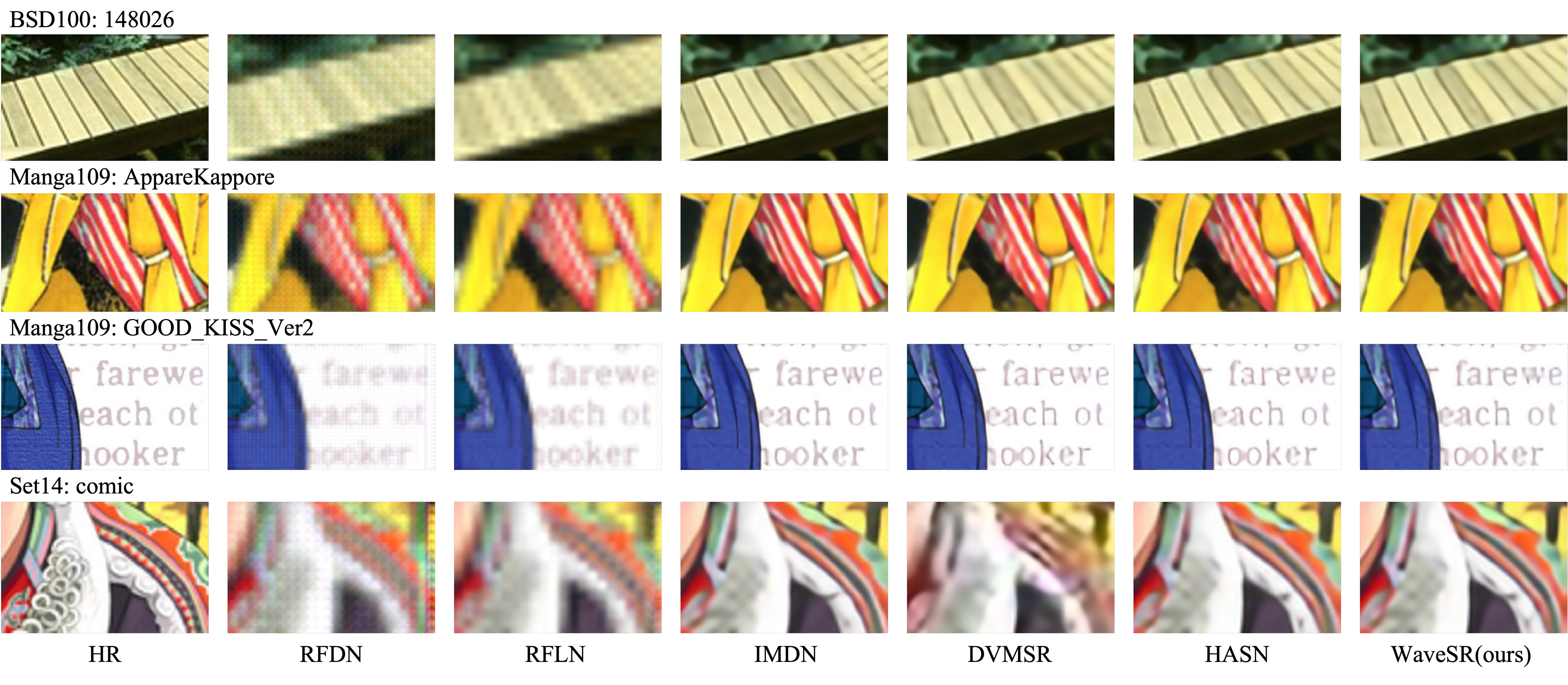}
\caption{Qualitative comparisons between existing efficient SR methods and our WaveSR.}
\label{fig:vis}
\end{figure}

\begin{figure}[!htbp]
\centering
\includegraphics[width=3.3in]{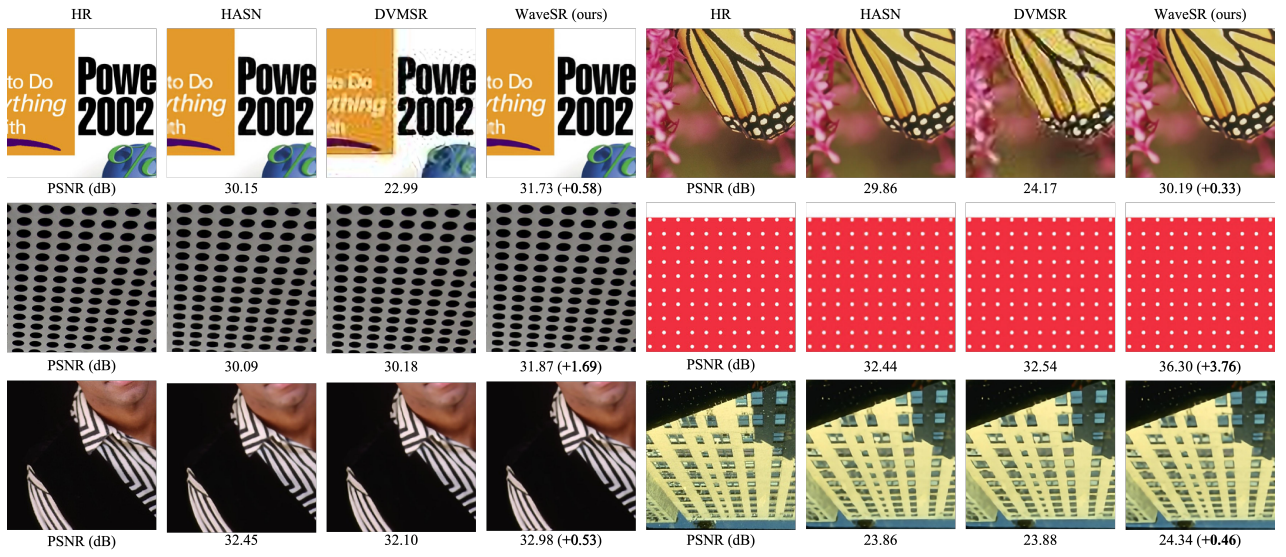}
\caption{Finding regions where wave evolution excels in image super-resolution. The values in parentheses indicate the minimum PSNR improvement compared to the other two methods.}
\label{fig:new_case}
\end{figure}

We compare our method with state-of-the-art ones and use the PSNR and SSIM to evaluate the quality of SR images. As shown in Table~\ref{tab:results}, WaveSR achieves competitive performance across the five benchmarks under a 414K/22.4G budget.
Visualization results in Fig.~\ref{fig:vis} demonstrate that WaveSR delivers superior reconstruction quality on images featuring architectural structures, fabric textures, English letters, and complex clothing backgrounds, highlighting the advantage of the wave equation solution in adaptive frequency separation and detail enhancement.
\\
\textbf{Further analysis.}
We compare two methods with similar model size and computational complexity to WaveSR, namely HASN and DVMSR (as shown in Fig.~\ref{fig:new_case}). By evaluating PSNR on representative regions from the test sets, including periodic grid/dot patterns and fine biological textures, the results demonstrate that WaveSR achieves superior reconstruction performance on high-frequency and structured texture images.
Notably, although these images are relatively simple in content, they contain both flat regions and high-frequency details, demonstrating the potential of wave evolution in handling such types of images.

\subsubsection{Comparison with Real-World SR Methods}

We compare WaveSR with several representative real-world SR methods using PSNR and LPIPS as evaluation metrics. All compared methods are GAN-based. Following DASR~\cite{dasr}, we retrain a real-world version of our model, denoted as WaveSR-Real. Quantitative results on four benchmark datasets are reported in Table~\ref{tab:real}, where ``Bicubic" denotes the DIV2K validation set downsampled via bicubic degradation in MATLAB.

\begin{table}[!htbp]
    \centering
    \caption{Quantitative comparison with representative real-world SR methods on benchmark datasets. The best results are highlighted in \textbf{bold}.}
    \resizebox{\linewidth}{!}{
    \begin{tabular}{llccccccc}
    \hline
    \multirow{2}{*}{Datasets} & \multirow{2}{*}{Metric}    & Real-ESRGAN  & SwinIR  & DASR  & HAT  & WaveSR-Real  \\
     &   &  \cite{realesrgan} &  \cite{swinir} &  \cite{dasr} & \cite{chen2023hat}  & (Ours) \\
    \hline
    \hline
    \multirow{2}{*}{Bicubic} & $\mathrm{PSNR}~(\uparrow)$    & 26.65 & 27.21 &\textbf{28.55} & 27.44 & 28.51\\
    & $\mathrm{LPIPS}~(\downarrow)$  &0.228 & 0.214& 0.170& 0.203 &\textbf{0.163}\\\hline

    \multirow{2}{*}{RealSR-cano~\cite{real-SR-cano}} & $\mathrm{PSNR}~(\uparrow)$  & 26.14 & 26.64 & \textbf{27.40} & 26.68 & 27.25 \\
    & $\mathrm{LPIPS}~(\downarrow)$  &0.378 &0.357 &0.393 &{0.342} &\textbf{0.280} \\\hline

    \multirow{2}{*}{RealSR-Nikon~\cite{real-SR-cano}} & $\mathrm{PSNR}~(\uparrow)$  &25.49 & 25.76& \textbf{26.35} & 25.85& 26.32\\
     & $\mathrm{LPIPS}~(\downarrow)$  & 0.388& 0.364 &0.401 &{0.358}  & \textbf{0.283} \\\hline
     
    \multirow{2}{*}{AIM2019-val~\cite{AIM2019-val}}  & $\mathrm{PSNR}~(\uparrow)$  &23.89 &23.89 & 23.76& \textbf{24.19}  &24.06\\
     & $\mathrm{LPIPS}~(\downarrow)$   &0.396 & 0.387&0.421 &{0.370}  &\textbf{0.283}\\\hline
    \hline
    \end{tabular}}
    \label{tab:real}
\end{table}

As shown in Table~\ref{tab:real}, WaveSR-Real achieves the lowest LPIPS across all four datasets, outperforming all competing methods by a clear margin. This demonstrates the superiority of our approach in terms of perceptual quality. In terms of PSNR, WaveSR-Real achieves competitive results, ranking second on Bicubic and RealSR-Nikon, while maintaining a small gap to the best-performing method on the other two datasets. These results indicate that WaveSR-Real strikes a favorable balance between perceptual quality and reconstruction fidelity, validating its effectiveness in real-world SR scenarios.

\section{Ablation Study}
In this section, we conduct a series of comparative experiments to demonstrate the effectiveness of our module design. 
\subsection{Understanding Wave Evolution in Image Super-Resolution}

\noindent\textbf{Importance of Frequency-Coordinate Correspondence.}
We ablate the physical angular-frequency map \(\omega\) to verify that WEA's performance stems from frequency-dependent propagation rather than a generic spectral transform. As shown in Table~\ref{tab:physics_omega}, randomly permuting \(\omega\) breaks the coordinate-coefficient alignment, and using a random constant \(\omega\) removes the frequency-dependent responses. The correctly ordered \(\omega\) achieves the best performance, with the largest margins observed on Urban100 and Manga109, datasets rich in repetitive structures and fine textures where wave propagation effects are most distinguishable. This indicates that WEA's effectiveness relies on the physically meaningful frequency-coordinate correspondence inherent to wave dynamics, rather than a generic spectral modulation strategy, validating the necessity of its wave-based formulation.

\begin{table}[!htbp]
    \centering
    \caption{Ablation on the frequency-coordinate correspondence in WEA for $\times4$ SR. The best results are highlighted in \textbf{bold}.}
    \resizebox{1\linewidth}{!}{
    \begin{tabular}{l|ccccccc}
        \toprule
        \multirow{2}{*}{Method} & Set5 & Set14 & BSD100 & Urban100 & Manga109 \\
    
         & PSNR/SSIM & PSNR/SSIM & PSNR/SSIM & PSNR/SSIM & PSNR/SSIM \\
        \midrule
        Correct $\omega$ &  \textbf{31.51}/\textbf{0.8861} & \textbf{28.17}/\textbf{0.7720} & \textbf{27.31}/\textbf{0.7285}  & \textbf{25.32}/\textbf{0.7593}  & \textbf{29.30}/\textbf{0.8922}  \\
        Permuted $\omega$ &  31.49/0.8861  & 28.14/0.7716   & 27.29/0.7280  & 25.28/0.7584  &  29.22/0.8908 \\
        Constant $\omega$ & 31.51/0.8858  &  28.15/0.7713  &  27.30/0.7275 & 25.29/0.7571  &  29.20/0.8905 \\
        \bottomrule
    \end{tabular}}
    \label{tab:physics_omega}
\end{table}

\begin{figure}[!htbp]
\centering
\includegraphics[width=3in]{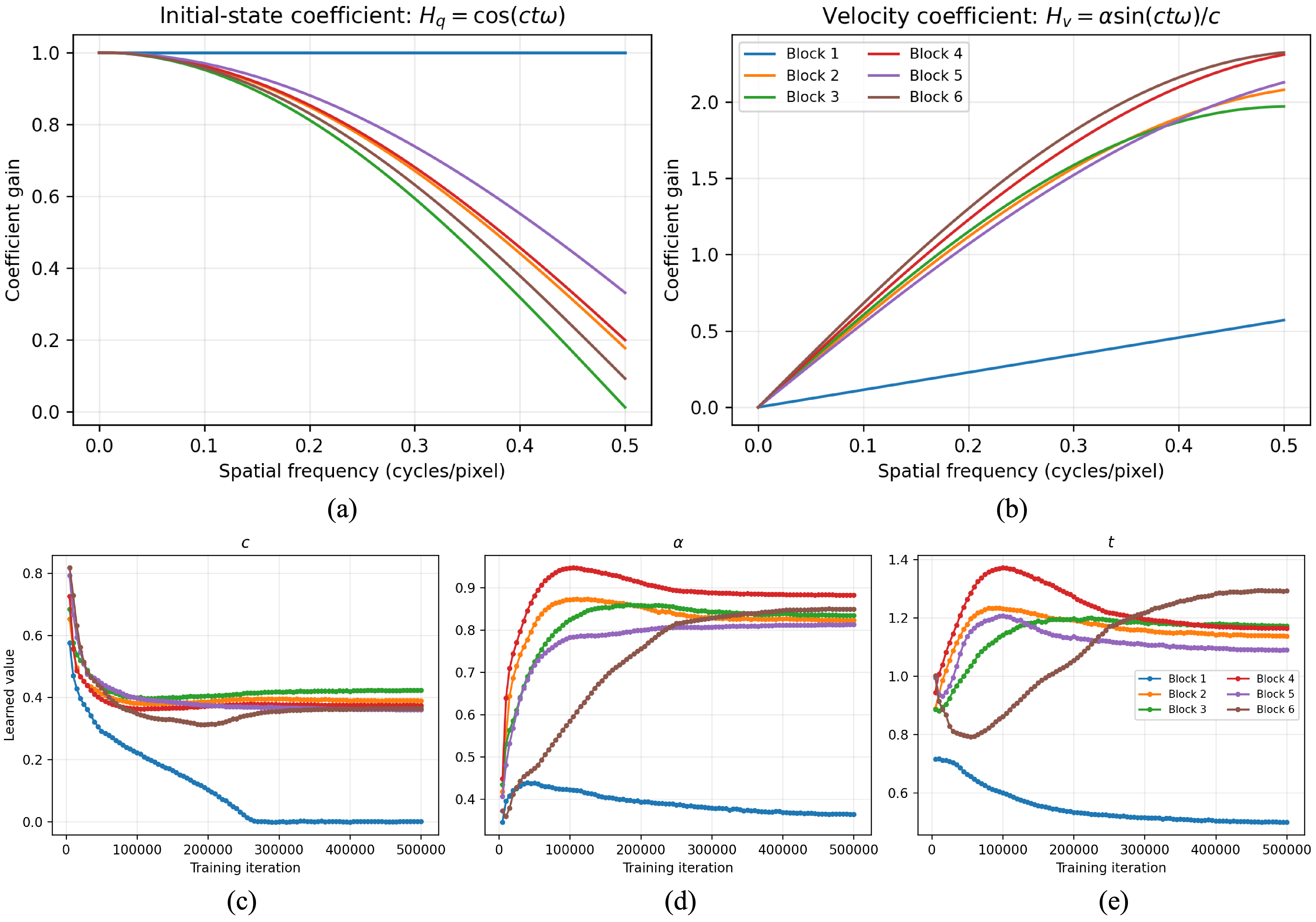}
\caption{Analysis of the frequency responses and learnable parameters in WEA. (a) Initial-state frequency responses $H_q(\omega)$. (b) Velocity-like frequency responses $H_v(\omega)$. (c)--(e) Training trajectories of the learnable wave-inspired parameters $c$, $\alpha$, and $t$, respectively, across different WEA blocks.}
\label{fig:wea_frequency_response}
\end{figure}

\noindent\textbf{Analysis of Frequency Responses and Learnable Parameters.}
As shown in Figs.~\ref{fig:wea_frequency_response}(a) and (b), different WEA blocks learn distinct frequency-dependent responses. The initial-state coefficient $H_q(\omega)$ mainly preserves low-frequency information, while the velocity-like coefficient $H_v(\omega)$ provides increasing modulation toward the middle- and high-frequency ranges. Figs.~\ref{fig:wea_frequency_response}(c)-(e) show that the wave-inspired parameters $c$, $\alpha$, and $t$ are actively optimized and converge to stable layer-dependent values. These results demonstrate that WEA learns structured spectral modulation across network depth, with distinct components specializing in different frequency ranges.

\noindent\textbf{Analysis of the Velocity-like Feature Assignment.}
To examine whether $K-Q$ provides a meaningful velocity-like representation, we compare it with zero velocity and two direct feature assignments, $K$ and $Q$. As shown in Table~\ref{tab:velocity_assignment}, setting the velocity-like component to zero consistently degrades performance, particularly on Urban100 and Manga109, indicating that the velocity-like branch contributes to the reconstruction of structurally complex content. Directly using $K$ or $Q$ provides limited improvements because these representations retain substantial static content and do not explicitly describe a feature-transition direction. In contrast, $K-Q$ achieves the best PSNR and SSIM on all five benchmarks. The larger gains on Urban100 and Manga109 suggest that the residual velocity-like feature is particularly useful for reconstructing edges, repeated structures, and dense textures.

\begin{table}[!htbp]
    \centering
    \caption{Ablation study of different velocity-like feature assignments in WEA. ``0'' denotes a zero velocity-like feature. The best results are highlighted in \textbf{bold}.}
    \resizebox{\linewidth}{!}{
    \begin{tabular}{l|ccccc}
        \toprule
        \multirow{2}{*}{\shortstack{Velocity-like\\feature $\mathbf{V}$}}
        & Set5 & Set14 & BSD100 & Urban100 & Manga109 \\
        & PSNR/SSIM & PSNR/SSIM & PSNR/SSIM
        & PSNR/SSIM & PSNR/SSIM \\
        \midrule
        $\mathbf{0}$
        & 31.49/0.8860
        & 28.11/0.7708
        & 27.27/0.7271
        & 25.23/0.7559
        & 29.09/0.8892 \\

        $\mathbf{K}$
        & 31.46/0.8856
        & 28.14/0.7711
        & 27.28/0.7273
        & 25.26/0.7567
        & 29.24/0.8905 \\

        $\mathbf{Q}$
        & 31.48/0.8860
        & 28.12/0.7713
        & 27.29/0.7277
        & 25.26/0.7572
        & 29.16/0.8904 \\

        $\mathbf{K}-\mathbf{Q}$
        & \textbf{31.51}/\textbf{0.8861}
        & \textbf{28.17}/\textbf{0.7720}
        & \textbf{27.31}/\textbf{0.7285}
        & \textbf{25.32}/\textbf{0.7593}
        & \textbf{29.30}/\textbf{0.8922} \\
        \bottomrule
    \end{tabular}}
    \label{tab:velocity_assignment}
\end{table}

\begin{figure}[!htbp]
\centering
\includegraphics[width=2.2in]{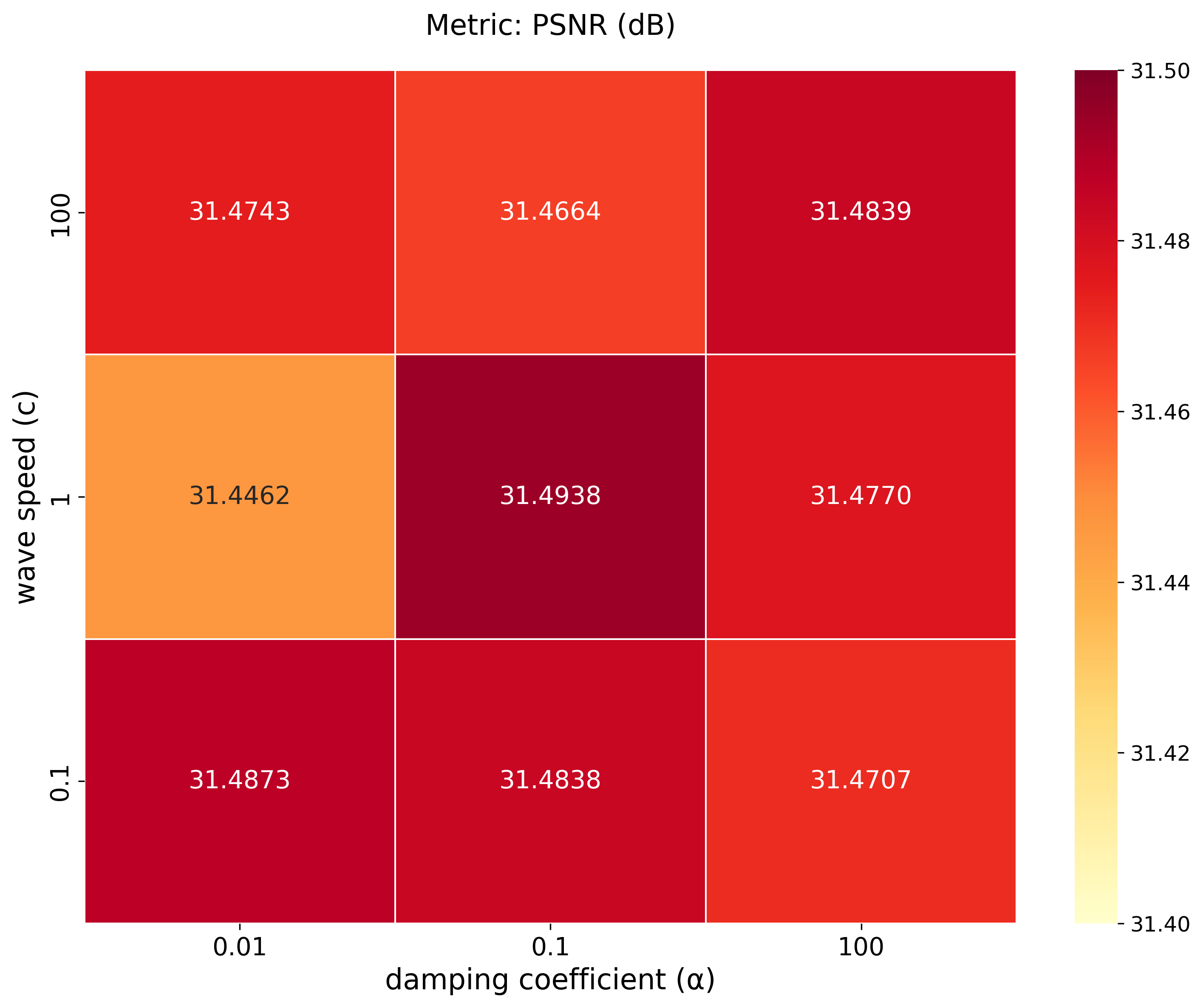}
\caption{Evaluation of thermal wave speed and damping term with different settings.}
\label{fig:ac}
\end{figure}

\noindent\textbf{Initialization Sensitivity of Wave Parameters.}
Although \(c\), \(\alpha\), and \(t\) are learnable, their initialization may affect optimization. We therefore evaluate different initial values \(c\) and \(\alpha\) while allowing both parameters to be updated during training. As shown in Fig.~\ref{fig:ac}, initialization with \(c=1\) and \(\alpha=0.1\) yields the highest PSNR on Set5 among the evaluated configurations. This result indicates that WEA exhibits moderate sensitivity to its initial parameter configuration.

\noindent\textbf{Learnability of Wave Parameters.}
Table~\ref{tab:physics} investigates whether the wave parameters \(c\) and \(\alpha\) should be learned from the training data. Fixing \(c\) has only a limited influence on the relatively simple Set5 dataset and even produces a slightly higher PSNR, likely because adapting the propagation behavior offers less benefit on low-frequency-dominated images and may introduce additional optimization difficulty. However, jointly learning \(c\) and \(\alpha\) provides more consistent improvements on datasets containing complex structures and abundant high-frequency details. These results suggest that the learnable damping parameter \(\alpha\) plays an important role in adapting the frequency attenuation behavior, while learning \(c\) further improves performance on datasets with more diverse spatial-frequency distributions. Overall, despite modest gains on simpler datasets, the results favor learning \(c\) and \(\alpha\) over using fixed values.
\begin{table}[!htbp]
    \centering
    \caption{Ablation study on the learnability of the wave parameters $c$ and $\alpha$ for $\times4$ SR. When a parameter is fixed, it is frozen at its initial value throughout training, while the remaining parameters are optimized normally. The best results are highlighted 
    in \textbf{bold}.}
    \label{tab:physics}
    \setlength{\tabcolsep}{5.0pt}
    \renewcommand{\arraystretch}{1.10}
    \resizebox{\linewidth}{!}{
    \begin{tabular}{l|ccccc}
        \toprule
        \multirow{2}{*}{Variant}
        & Set5 & Set14 & BSD100 & Urban100 & Manga109 \\
        & PSNR/SSIM & PSNR/SSIM & PSNR/SSIM
        & PSNR/SSIM & PSNR/SSIM \\
        \midrule
        Fixed $c$ = 1
        & \textbf{31.54}/\textbf{0.8864}
        & \textbf{28.17}/0.7718
        & \textbf{27.31}/0.7279
        & 25.31/0.7582
        & 29.24/0.8913 \\

        Fixed $\alpha$ = 0.1
        & 31.52/0.8862
        & 28.15/0.7715
        & 27.29/0.7277
        & 25.27/0.7571
        & 29.17/0.8902 \\

        Learnable $c$ and $\alpha$ 
        & 31.51/0.8861
        & \textbf{28.17}/\textbf{0.7720}
        & \textbf{27.31}/\textbf{0.7285}
        & \textbf{25.32}/\textbf{0.7593}
        & \textbf{29.30}/\textbf{0.8922} \\
        \bottomrule
    \end{tabular}}
\end{table}

\subsection{Effectiveness of Wave Equation Solution}
\begin{figure}[!htbp]
    \centering
    \includegraphics[width=3in]{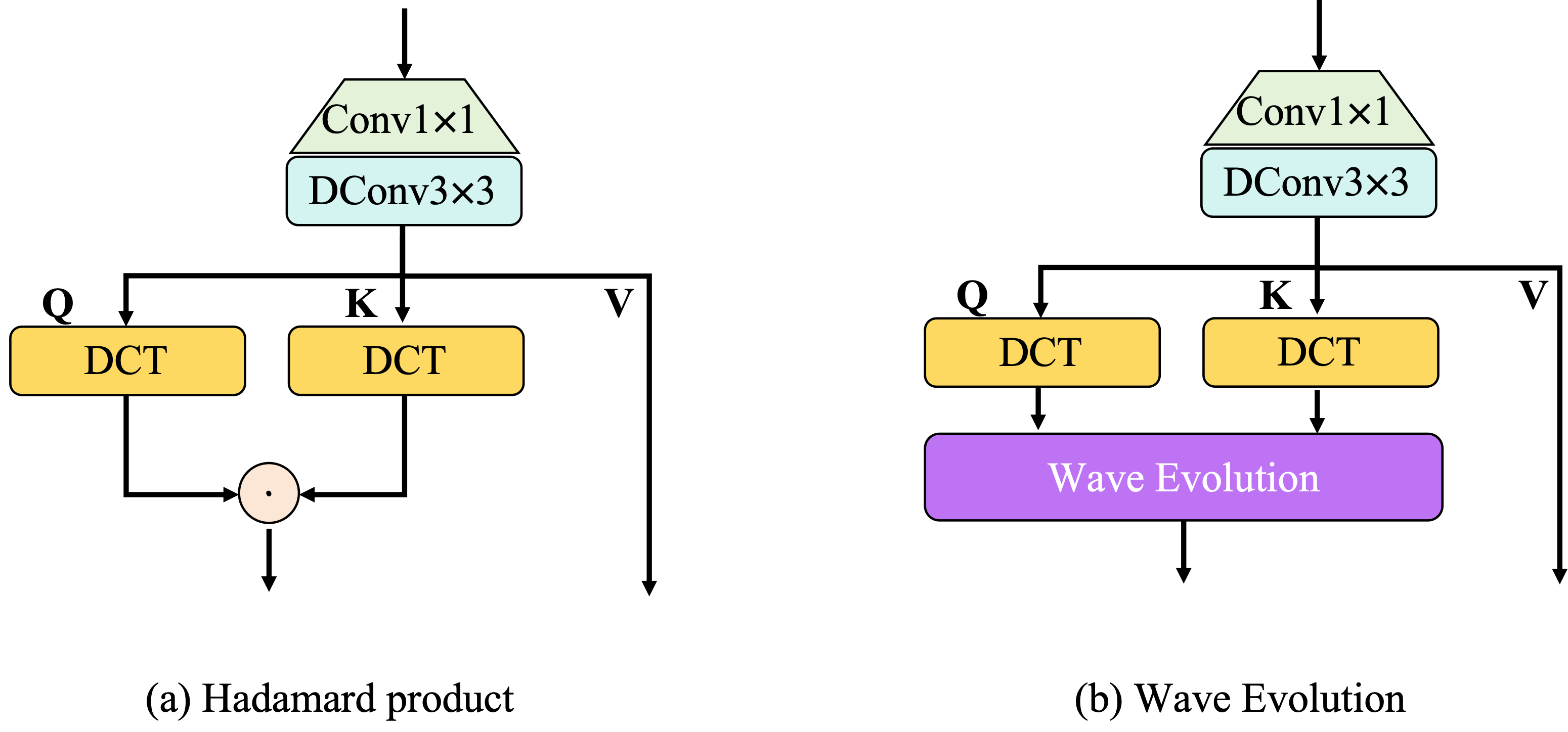}
    \caption{Architectural comparison of two feature interaction
    schemes: (a) direct Hadamard-product interaction and
    (b) the proposed wave-equation-guided evolution. The remaining network components are kept identical.}
    \label{fig:ab1}
\end{figure}

\noindent\textbf{Comparison with Direct Hadamard Interaction.}
We investigate whether the proposed wave-equation-guided evolution provides a more effective feature interaction mechanism than a direct Hadamard product. As illustrated in Fig.~\ref{fig:ab1}, we construct two networks that share the same backbone and differ only in their feature interaction schemes. The first network directly performs element-wise Hadamard interaction, whereas the second adopts the proposed wave-equation-guided evolution. All other network components and training settings are kept unchanged.

\begin{figure}[!htbp]
\centering
\includegraphics[width=3in]{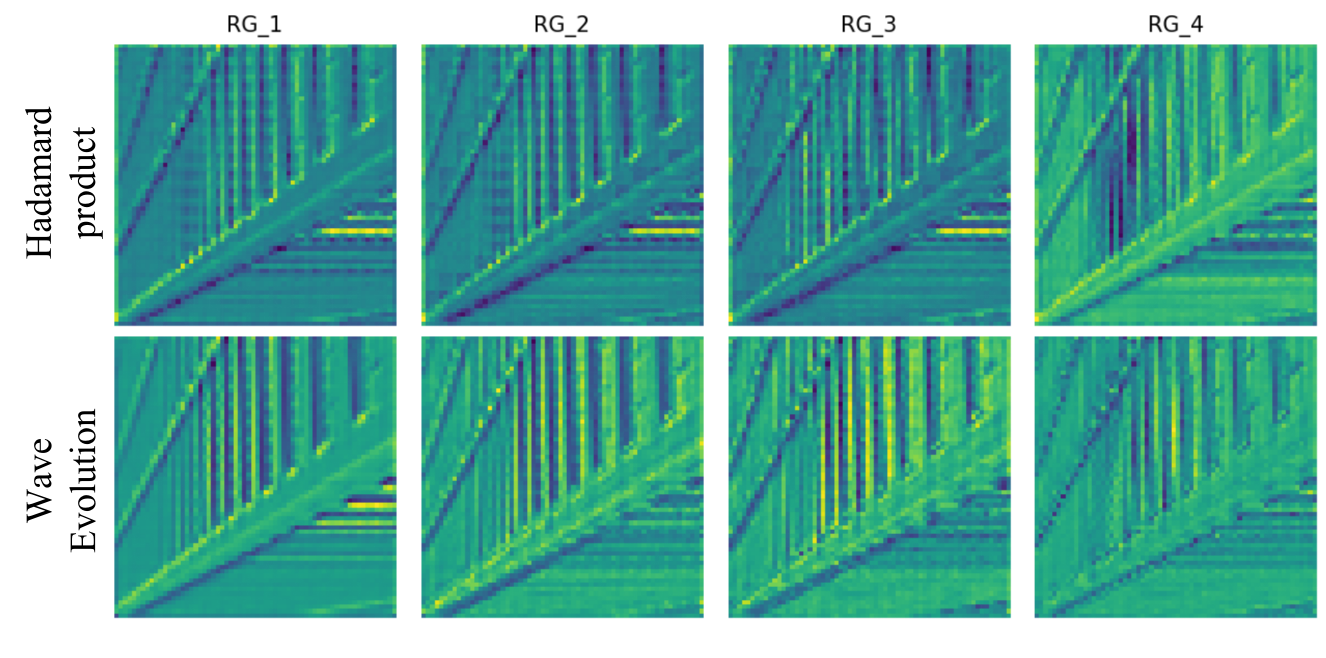}
\caption{The spectral feature maps of two feature interaction schemes.}
\label{fig:ab_wave}
\end{figure}

As reported in Table~\ref{tab:hadamard_comparison}, these results demonstrate that the reconstruction improvements are achieved without increasing model capacity or computational overhead, indicating that the structured wave-equation-guided interaction is more effective than direct element-wise feature modulation. To provide further insight into the internal feature behavior, we visualize the layer-wise spectral feature maps generated by the two interaction schemes in Fig.~\ref{fig:ab_wave}. The spectral responses of the Hadamard-product interaction are mainly concentrated in the low- and middle-frequency bands, while its high-frequency responses become progressively weaker in deeper residual groups. In contrast, the proposed wave-equation-guided evolution retains responses over a broader frequency range and exhibits relatively less attenuation of high-frequency components across different residual groups.
\begin{table}[!htbp]
    \centering
    \caption{Controlled comparison between direct Hadamard-product interaction and the proposed wave-equation-guided evolution. Inference time and max memory are measured under identical settings. The best results are highlighted in \textbf{bold}.}
    \resizebox{1\linewidth}{!}{
    \begin{tabular}{l|cccccccc}
        \toprule
        \multirow{2}{*}{Method} & \multirow{2}{*}{\#Params[K]} & \multirow{2}{*}{FLOPs[G]} & \multirow{2}{*}{Time[ms]} & \multirow{2}{*}{Max Memory[M]} & Set5 & Set14\\
         & & & & & PSNR/SSIM & PSNR/SSIM \\
        \midrule
        Hadamard Product & 183 &  12.35 & 82.98 & 1191  & 30.94/0.8760& 27.79/0.7630\\
        Wave Evolution & \textbf{177} & \textbf{12.18}& \textbf{80.41} &  \textbf{1078}&\textbf{31.34}/\textbf{0.8836} &	\textbf{28.06}/\textbf{0.7693}\\
        \bottomrule
    \end{tabular}}
    \label{tab:hadamard_comparison}
\end{table}

\begin{table*}[!t]
    \centering
    \caption{Controlled comparison of WEA with generic Fourier-based variants on Urban100 and Manga109. Lower values are better for FFT-L1 and Laplacian-L1, while higher values are better for PSNR and SSIM. The best reconstruction results are highlighted in \textbf{bold}.}
    \label{tab:wave_equation_analysis}
    
    \renewcommand{\arraystretch}{1.10}
    \setlength{\tabcolsep}{4.5pt}
    
    \resizebox{0.9\textwidth}{!}{%
    \begin{tabular}{llccccccccc}
        \toprule
        \multirow{2}{*}{Dataset}
        & \multirow{2}{*}{Variant}
        & \multirow{2}{*}{\#Params [K]}
        & \multirow{2}{*}{FLOPs [G]}
        & \multirow{2}{*}{PSNR/SSIM}
        & \multicolumn{3}{c}{FFT-L1~$(\downarrow)$}
        & \multirow{2}{*}{Laplacian-L1~$(\downarrow)$}
        & \multicolumn{2}{c}{Region-wise PSNR~$(\uparrow)$} \\
        \cmidrule(lr){6-8}
        \cmidrule(lr){10-11}
        & & & & 
        & Low & Mid & High
        & & Edge & Texture \\
        \midrule

        \multirow{4}{*}{Urban100}
        & FFT-Mixer
        & 177 & 11.65
        & 24.86/0.7417
        & 0.0765 & 0.0521 & 0.0259
        & 0.3456
        & 18.8291 & 23.9925 \\

        & Frequency Gate
        & 199 & 11.65
        & 25.26/0.7566
        & \textbf{0.0720} & 0.0510 & \textbf{0.0258}
        & 0.3351
        & 19.2549 & 24.3244 \\

        & Complex Filter
        & 220 & 11.65
        & 25.28/0.7574
        & 0.0726 & 0.0512 & \textbf{0.0258}
        & 0.3365
        & 19.1799 & 24.2695 \\

        & WEA
        & 177 & 11.65
        & \textbf{25.32/0.7593}
        & \textbf{0.0720}
        & \textbf{0.0508}
        & \textbf{0.0258}
        & \textbf{0.3328}
        & \textbf{19.3244}
        & \textbf{24.3361} \\

        \midrule

        \multirow{4}{*}{Manga109}
        & FFT-Mixer
        & 177 & 11.65
        & 28.23/0.8738
        & 0.0681 & 0.0311 & 0.0095
        & 0.1733
        & 21.3006 & 29.3825 \\

        & Frequency Gate
        & 199 & 11.65
        & 29.14/0.8899
        & 0.0604 & 0.0293 & \textbf{0.0094}
        & 0.1597
        & 22.2309 & 30.1813 \\

        & Complex Filter
        & 220 & 11.65
        & 29.15/0.8898
        & 0.0614 & 0.0296 & 0.0095
        & 0.1617
        & 22.0819 & 30.0459 \\

        & WEA
        & 177 & 11.65
        & \textbf{29.30/0.8922}
        & \textbf{0.0597}
        & \textbf{0.0290}
        & 0.0095
        & \textbf{0.1575}
        & \textbf{22.3841}
        & \textbf{30.2813} \\
        
        \bottomrule
    \end{tabular}}
\end{table*}
\noindent\textbf{Selection of Frequency-domain Transformation.} 
To examine the effect of different frequency-domain transformations in our wave evolution attention mechanism, comparing FFT with the widely used Discrete Cosine Transform (DCT). As shown in Table~\ref{tab:dctfft}, FFT achieves slightly but consistently better reconstruction accuracy than DCT on the evaluated benchmarks, while using the same number of parameters and maintaining comparable FLOPs and memory consumption.
Quantitatively, FFT improves PSNR/SSIM on both Set5 and Set14. This suggests that the complex spectral representation of FFT may be more suitable for modeling frequency-dependent wave evolution, especially for modulating high-frequency details that are important for super-resolution. However, we also observe that DCT has lower inference latency than FFT. Therefore, it show that FFT provides a small accuracy gain at the cost of a modest increase in runtime.

\begin{table}[!htbp]
    \centering
    \caption{Selection of Frequency-domain Transformation. The best results are highlighted in \textbf{bold}.}
    \resizebox{1\linewidth}{!}{
    \begin{tabular}{l|cccccccc}
        \toprule
        \multirow{2}{*}{Method} & \multirow{2}{*}{\#Params[K]} & \multirow{2}{*}{FLOPs[G]} & \multirow{2}{*}{Time[ms]} & \multirow{2}{*}{Max Memory[M]} & Set5 & Set14\\
         & & & & & PSNR/SSIM & PSNR/SSIM \\
        \midrule
        DCT & 414 & 23.5& \textbf{99.7} & 2165.7 & 31.89/0.8917 &	28.40/0.7781\\
        FFT & 414 & \textbf{22.4} & 118.7 & \textbf{2163.6} & \textbf{31.97}/\textbf{0.8927} &	\textbf{28.41}/\textbf{0.7785}\\
        
        \bottomrule
    \end{tabular}}
    \label{tab:dctfft}
\end{table}

\noindent\textbf{Controlled Comparison with Generic Fourier-Based Variants.}
To distinguish WEA from generic frequency-domain processing, we compare it with FFT-Mixer, Frequency Gate, and Complex Filter. As shown in Table~\ref{tab:wave_equation_analysis}, WEA achieves the best performance on both datasets with only 177K parameters and 11.65G FLOPs. It outperforms the strongest alternative by 0.04/0.15~dB on Urban100/Manga109 and the parameter-matched FFT-Mixer by 0.46/1.07~dB, indicating that the gains are not merely due to model capacity. WEA also yields the lowest mid-frequency and Laplacian errors and the highest Edge-PSNR and Texture-PSNR on both datasets. These results suggest that its wave-equation-constrained modulation achieves more balanced frequency modeling and better recovers structural edges and fine textures.

\noindent\textbf{Comparison with Structurally Related FFN Variants.}
We further compare FA-FFN with structurally related spatial- and Fourier-domain FFN variants to assess its efficiency and effectiveness. As shown in Table~\ref{tab:ffn}, FA-FFN achieves the best PSNR and SSIM on both Urban100 and Manga109, which contain abundant repetitive structures and fine textures. Notably, FA-FFN accomplishes this with the fewest parameters among all compared methods and maintains competitive computational cost with 14.37G FLOPs and 74ms inference time. Compared to GFNet, FA-FFN reduces parameters by 19.6\% while delivering superior reconstruction quality. Although FA-FFN uses slightly more memory than the baseline FFN, this overhead is marginal relative to its performance gains. These results demonstrate that the proposed frequency-aware feature interaction effectively captures high-frequency details without introducing excessive computational burden.

\begin{table}[!htbp]
    \centering
    \caption{Comparison with Structurally Related FFN Variants. All variants are evaluated using the same backbone, training protocol, and inference setting. The best results are highlighted in \textbf{bold}.}
    \resizebox{1\linewidth}{!}{
    \begin{tabular}{l|cccc|cc}
        \toprule
        \multirow{2}{*}{Method} & \multirow{2}{*}{\#Params[K]} & \multirow{2}{*}{FLOPs[G]} & \multirow{2}{*}{Time[ms]} & \multirow{2}{*}{Max Memory[M]}  & Urban100 & Manga109 \\
         & & & & & PSNR/SSIM & PSNR/SSIM\\
        \midrule
        FFN~\cite{vaswani2017attention} & 301  & 19.53 & \textbf{46}  & \textbf{999}  & 24.96/0.7423 & 28.72/0.8791 \\
         GFNet~\cite{rao2021global} & 362 &  14.37 &  75   &  2101  & 25.64/{0.7694} & 29.93/{0.8996} \\
        AFNO~\cite{guibas2021adaptive} & 355 & 19.75  &   96 & 2471 &  25.47/0.7633 & 29.67/0.8954 \\
        FA-FFN & \textbf{291} &  \textbf{14.37} &  {74}  &  2101  & \textbf{25.66}/\textbf{0.7697} & \textbf{29.94}/\textbf{0.8996}	\\
        \bottomrule
    \end{tabular}}
    \label{tab:ffn}
\end{table}

\subsection{The Architecture Design of WaveSR}
\begin{figure}[!htbp]
\centering
\includegraphics[width=3in]{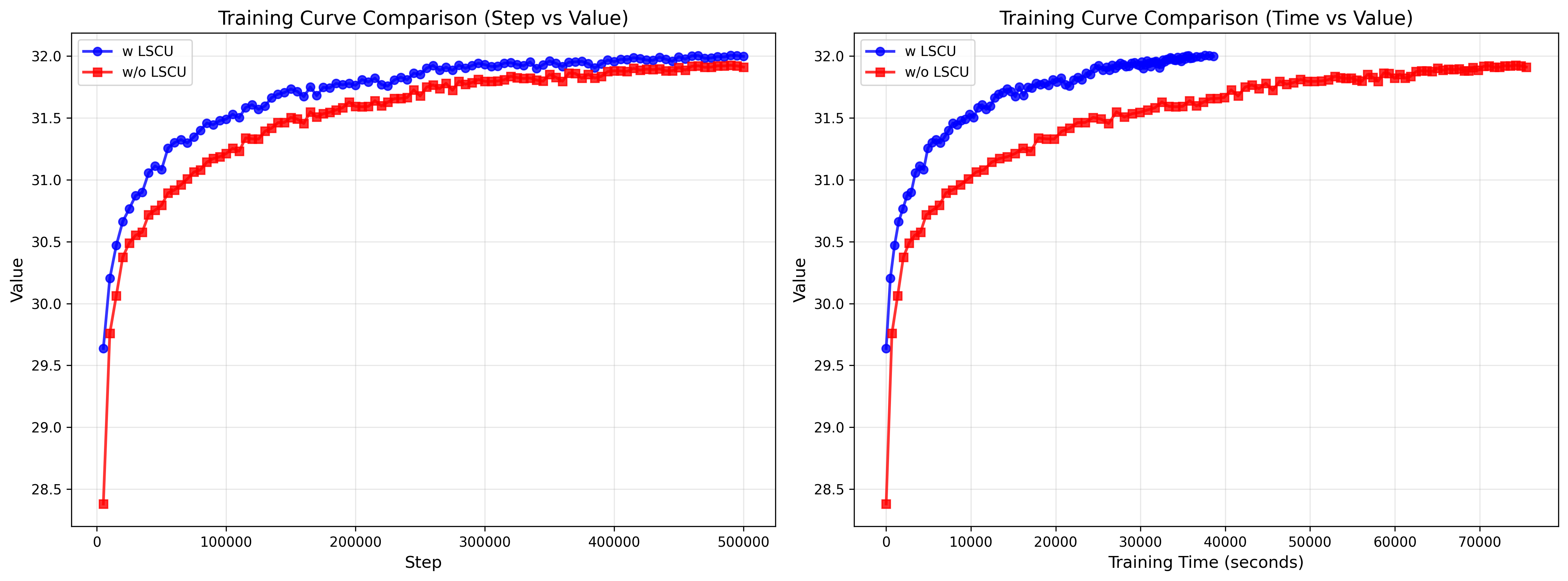}
\caption{Training performance comparison of models with/without Long Skip Connection with Upsampling (LSCU).}
\label{ab:curve}
\end{figure}

\noindent\textbf{Effectiveness of Long Skip Connection with Upsampling.} To investigate the effectiveness of Long Skip Connection with Upsampling, we conduct an ablation study comparing models with and without this connection, as shown in Fig.~\ref{ab:curve}. The left and right subfigures illustrate the PSNR values on Set5 as a function of training steps and training time, respectively. The results demonstrate that incorporating the Long Skip Connection with Upsampling enables the network to focus more on learning high-frequency details, leading to more stable training and faster convergence.

\noindent\textbf{Effectiveness of FA-FFN and WEA.}
As shown in Table~\ref{tab:abffn}, comparing the two configurations with only the standard FFN and only FA-FFN, FA-FFN achieves PSNR gains of 0.79 dB on Set5 and 0.52 dB on Set14, verifying its effectiveness as a frequency-aware feature refinement component. Introducing WEA into the standard FFN baseline improves performance by 0.72 dB and 0.44 dB, respectively, while replacing the standard FFN with FA-FFN yields further improvements of 0.20 dB and 0.15 dB. This indicates that WEA and FA-FFN operate on different aspects of feature modeling: WEA performs cross-feature interaction through structured frequency-dependent evolution, whereas FA-FFN adaptively refines individual spectral components. Their combination achieves the best performance, confirming their complementarity.

\begin{table}[!htbp]
    \centering
    \caption{Quantitative evaluations of each component in the proposed method. The best results are highlighted in \textbf{bold}.}
    \resizebox{0.8\linewidth}{!}{
    \begin{tabular}{ccccccc}
        \toprule
        \multirow{2}{*}{FFN~\cite{vaswani2017attention}} & \multirow{2}{*}{FA-FFN} & \multirow{2}{*}{WEA}  & Set5 & Set14\\
         & &   & PSNR/SSIM & PSNR/SSIM \\
        \midrule
        \ding{51}& \ding{56}&\ding{56} & 31.08/0.8773&	27.89/0.7642\\
        \ding{56} & \ding{51} & \ding{56} &31.87/0.8910 &28.41/0.7782	\\

        \ding{51} &  \ding{56}  & \ding{51}&   31.80/0.8900 & 28.33/0.7756	\\

         \ding{56} &  \ding{51}  & \ding{51}&   \textbf{32.00}/\textbf{0.8929} & \textbf{28.48}/\textbf{0.7795}	\\
        
        \bottomrule
    \end{tabular}}
    \label{tab:abffn}
\end{table}

\subsection{Study on Loss Function}
To investigate the effects of loss-function composition, the relative weighting between pixel-domain and frequency-domain losses, and the weight allocation across frequency bands, we train all models from scratch on the same dataset using an identical network architecture and a fixed training budget of 500K iterations.
\begin{table}[!htbp] 
\centering
\caption{Quantitative comparison of different loss-function configurations on Urban100 and Manga109. The frequency weights are listed in the order ($w_L:w_M:w_H$). Smooth, Edge, and Texture denote the corresponding region-wise PSNR values. The best results are highlighted in \textbf{bold}.}
\label{tab:abloss}
\setlength{\tabcolsep}{3.2pt}
\renewcommand{\arraystretch}{1.10}
\resizebox{0.45\textwidth}{!}{
\begin{tabular}{cccc|rrrr|rrrr}
\toprule
\multirow{2}{*}{$(\mathcal{L}_{1})$}
& \multirow{2}{*}{$(\mathcal{L}_{\mathrm{freq}})$}
& \multirow{2}{*}{$(\lambda_{1}:\lambda_{\mathrm{freq}})$}
& \multirow{2}{*}{$(w_L:w_M:w_H)$}
& \multicolumn{4}{c|}{Urban100}
& \multicolumn{4}{c}{Manga109} \\
\cmidrule(lr){5-8}\cmidrule(lr){9-12}
& & & &
PSNR/SSIM & Smooth & Edge & Texture &
PSNR/SSIM & Smooth & Edge & Texture \\
\midrule
\ding{51} & \ding{56} & -- & -- &
25.75/0.7752 & 32.27 & 19.79 & 24.68 &
29.99/0.9021 & \textbf{39.83} & 23.08 & 30.92 \\

\ding{56} & \ding{51} & -- & 1:1:1 &
25.80/0.7747 & 32.27 & 19.86 & 24.73 &
30.19/0.9030 & 39.61 & 23.34 & 30.93 \\

\ding{51} & \ding{51} & 1:1 & 1:1:1 &
\textbf{25.84}/\textbf{0.7757} &
\textbf{32.30} & \textbf{19.89} & \textbf{24.74} &
\textbf{30.21}/\textbf{0.9035} &
39.54 & \textbf{23.35} & \textbf{30.95} \\
\midrule
\ding{51} & \ding{51} & 1:1 & 2:1:1 &
25.79/0.7744 & 32.28 & 19.84 & 24.72 &
30.16/0.9030 & 39.50 & 23.31 & 30.91 \\

\ding{51} & \ding{51} & 1:1 & 1:2:1 &
25.79/0.7740 & 32.25 & 19.85 & 24.70 &
30.14/0.9025 & 39.43 & 23.29 & 30.88 \\

\ding{51} & \ding{51} & 1:1 & 1:1:2 &
25.78/0.7744 & 32.27 & 19.83 & 24.72 &
30.20/0.9031 & 39.60 & 23.35 & 30.95 \\
\midrule
\ding{51} & \ding{51} & 1:1 & 1:0:0 &
22.77/0.6406 & 29.09 & 16.83 & 21.84 &
24.31/0.7709 & 33.42 & 17.53 & 25.00 \\

\ding{51} & \ding{51} & 1:1 & 0:1:0 &
24.56/0.7200 & 30.46 & 18.72 & 23.45 &
27.52/0.8522 & 35.02 & 20.95 & 27.65 \\

\ding{51} & \ding{51} & 1:1 & 0:0:1 &
25.00/0.7366 & 30.40 & 19.20 & 23.84 &
28.32/0.8441 & 34.32 & 21.89 & 28.32 \\
\midrule
\ding{51} & \ding{51} & 2:1 & 1:1:1 &
25.80/0.7746 & 32.27 & 19.86 & 24.73 &
30.17/0.9027 & 39.43 & 23.32 & 30.90 \\

\ding{51} & \ding{51} & 1:2 & 1:1:1 &
25.81/0.7747 & 32.27 & 19.87 & 24.72 &
30.18/0.9030 & 39.53 & 23.33 & 30.92 \\
\bottomrule
\end{tabular}}
\end{table}
With the balanced frequency weighting $(w_L:w_M:w_H=1:1:1)$, using the frequency-domain loss alone improves PSNR over the $\mathcal{L}_{1}$-only baseline by 0.05~dB on Urban100 and 0.20~dB on Manga109. Combining the two losses with $(\lambda_{1}:\lambda_{\mathrm{freq}}=1:1)$ and the same balanced weighting achieves the best overall PSNR/SSIM on both datasets, outperforming the $\mathcal{L}_{1}$-only baseline by 0.09~dB and 0.22~dB, respectively.
Region-wise analysis reveals that the combined loss consistently enhances Edge and Texture regions. For Smooth regions, the effect is dataset-dependent: a slight improvement of 0.03~dB is observed on Urban100, while a minor decrease of 0.29~dB occurs on Manga109. Notably, the Smooth PSNR on Manga109 remains at a high level of 39.54~dB even with this decrease, and the overall PSNR/SSIM still shows a clear gain over the baseline.
Moderately re-weighting the frequency bands with (2:1:1), (1:2:1), or (1:1:2) yields only marginal and inconsistent variations, whereas any single-frequency supervision alone causes a marked performance drop across nearly all metrics. These results confirm that the three frequency regions provide complementary supervisory signals, and that a balanced multi-frequency weighting leads to the most stable and robust reconstruction performance.

\subsection{Effect of Training Strategy}
To validate the effectiveness of our training strategy, we compare training from scratch with our proposed progressive training strategy under identical model settings. As shown in Table~\ref{tab:trainingstrategy}, progressive training consistently outperforms training from scratch across all five benchmarks, with marginal but stable gains in both PSNR and SSIM. This improvement demonstrates that progressive training better stabilizes the optimization of our physics-inspired wave evolution module, facilitating more effective learning of high-frequency details and leading to superior super-resolution performance without additional computational overhead. 
\begin{table}[!htbp]
    \centering
    \caption{Effectiveness Study of the proposed training strategy. The best results are highlighted in \textbf{bold}.}
    \resizebox{1\linewidth}{!}{
    \begin{tabular}{c|ccccc}
        \toprule
        \multirow{2}{*}{Strategy} & Set5 & Set14 & BSD100& Urban100 & Manga100\\
         &PSNR/SSIM & PSNR/SSIM & PSNR/SSIM & PSNR/SSIM & PSNR/SSIM \\
        \midrule

        Scratch  & 32.12/0.8950 & 28.60/0.7823 & 27.59/0.7379 & 26.03/0.7814  & 30.54/0.9075	\\
         Progressive Training &  \textbf{32.21}/\textbf{0.8958}& \textbf{28.63}/\textbf{0.7829} & \textbf{27.62}/\textbf{0.7388} & \textbf{26.10}/\textbf{0.7845}  & \textbf{30.64}/\textbf{0.9086}	\\
        
        \bottomrule
    \end{tabular}}
    \label{tab:trainingstrategy}
\end{table}

\section{Conclusion}
In this paper, we propose WaveSR, a physics-inspired wave evolution attention for efficient image super-resolution. Specifically, it models the propagation and enhancement of high-frequency information in the frequency domain through the Wave Evolution Attention (WEA). Additionally, we introduce a Frequency-Adaptive Feed-Forward Network (FA-FFN), which enables adaptive frequency-aware feature selection through learnable spectral modulation, effectively enhancing feature representation. These components provide an interesting perspective for addressing lightweight image super-resolution. We believe that integrating fundamental physical principles into neural network design offers a promising direction for developing interpretable, efficient, and high-performance models.

\bibliographystyle{IEEEtran}
\bibliography{hifiwave}

\end{document}